\pdfoutput=1

\documentclass[11pt]{article}

\usepackage{ACL}

\usepackage{times}
\usepackage{latexsym}

\usepackage[T1]{fontenc}

\usepackage[utf8]{inputenc}
\usepackage{booktabs, multirow, multicol}
\usepackage[usestackEOL]{stackengine}
\usepackage{graphicx}	
\usepackage{color, colortbl}
\definecolor{Gray}{gray}{0.9}
\usepackage{subcaption}

\usepackage{stfloats}

\usepackage{microtype}

\usepackage{inconsolata}
\usepackage{amsmath}
\usepackage{cleveref}

\usepackage{enumitem}
\usepackage{makecell}

\title{Mind Your Format: Towards Consistent Evaluation\\of In-Context Learning Improvements}

\author{Anton Voronov \\
  Yandex, HSE University, MIPT \\
  \texttt{voronov.ad@phystech.edu} \\\And
  Lena Wolf\\
  Yandex, HSE University \\
  \texttt{e.a.volf@yandex.ru} \\\And
  Max Ryabinin \\
  Together AI \\
  \texttt{mryabinin0@gmail.com}}

\begin{document}
\newcommand{\nmodels}{19~}

\maketitle
\begin{abstract}
Large language models demonstrate a remarkable capability for learning to solve new tasks from a few examples.
The \textit{prompt template}, or the way the input examples are formatted to obtain the prompt, is an important yet often overlooked aspect of in-context learning.
In this work, we conduct a comprehensive study of the template format's influence on the in-context learning performance.
We evaluate the impact of the prompt template across 21 models (from 770M to 70B parameters) and 4 standard classification datasets. 
We show that a poor choice of the template can reduce the performance of the strongest models and inference methods to a random guess level.
More importantly, the best templates do not transfer between different setups and even between models of the same family.
Our findings show that the currently prevalent approach to evaluation, which ignores template selection, may give misleading results due to different templates in different works.
As a first step towards mitigating this issue, we propose \textit{Template Ensembles} that aggregate model predictions across several templates.
This simple test-time augmentation boosts average performance while being robust to the choice of random set of templates.
\end{abstract}

\section{Introduction}

Pretrained language models have emerged as a dominant paradigm for solving many NLP problems in a unified framework~\cite{gpt3,chowdhery2022palm,bloom,llama}.
In particular, these models can achieve impressive downstream results with just a few demonstrations given as a part of their input~\cite{good_examples,min-etal-2022-metaicl}, which is often called \textit{a prompt} in this case.
\vspace{8pt}

These few-shot or \textit{in-context learning} (ICL) abilities~\cite{gpt3} of large models are a subject of frequent study, as the primary factors behind them are not yet fully understood.
For example, one line of work investigates in-context learning within different theoretical frameworks~\cite{xie2022an,garg2022what,akurek2023what}.
In addition, multiple publications study the importance of different prompt attributes, such as the order of input demonstrations~\citep{lu-etal-2022-fantastically} and their labels~\citep{min-etal-2022-rethinking}.

\begin{figure*}[h!]
\includegraphics[width=\textwidth]{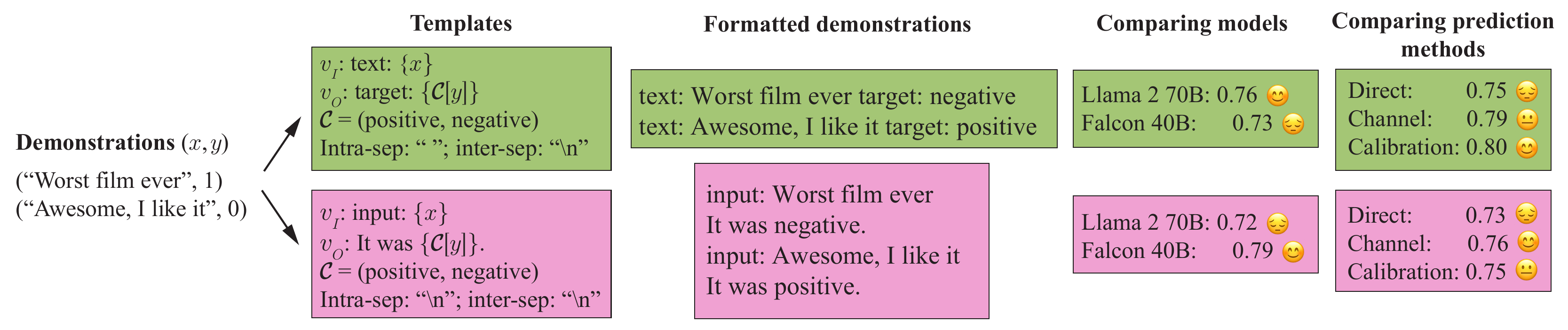}
\caption{An example template transformation for two demonstrations. Different prompt formats lead to different rankings both for models and ICL methods, and the best template for one method can be suboptimal for others.}
\label{figure:template_transform}
\end{figure*}

As shown in~\citet{calibrate,min-etal-2022-noisy}, the prompt format (i.e., a transformation from a set of examples to a natural language input) is also highly important.
However, this aspect is often overlooked in most existing studies.
Namely, works proposing modifications of ICL frequently present their results for a \textit{specific} template without specifying the criteria guiding its selection.
Furthermore, even when the results are averaged over a set of templates, they are compared to methods that were evaluated on a \textit{different} set of templates.
We illustrate this common discrepancy in~\Cref{app:prior_templates}.
Such inconsistency can lead to a misinterpretation of the reported results: the difference between the performance of two methods may be explained by the variation across prompt formats rather than the methods themselves.

In this work, we evaluate the template sensitivity of 21 models from 8 families,
including state-of-the-art open-access models, such as Llama 2~\cite{llama2} and Falcon~\cite{falcon}, as well as latest instruction-tuned models, such as Mistral~\cite{mistral} and Llama 3 Instruct~\cite{llama3}. 
We show that this issue persists regardless of the model size and the number of demonstrations.
Moreover, comparing various in-context learning enhancements while taking the template influence into account renders the superiority of one method over others less apparent.
Therefore, it is likely that the gains reported for advanced prompting methods can often be attributed to a luckily chosen template.
\pagebreak[4]

Crucially, \textbf{there are no universally best templates} for a given task. 
The best performing demonstration format for a fixed evaluation setting (i.e., the dataset, the model, the demonstration set, and the prediction method) does not transfer consistently across models (even within the same family), demonstration sets, or different prediction methods.
We find this concerning, as even the best template for a given setting can produce poor results after slight changes, which makes ``tuning'' the template a very difficult task.

As a first step towards addressing template sensitivity in a practical way, we propose \textbf{Template Ensembles} --- a test-time augmentation approach that averages model predictions over several prompt formats.
This method is easy to implement and increases the average performance across templates for multiple prompting methods while reducing the sensitivity of these methods to the template choice.

In summary, our contributions are as follows:
\begin{enumerate}
    \item We conduct a broad evaluation\footnote{Our code and results of all evaluations can be found at \href{https://github.com/yandex-research/mind-your-format}{\texttt{github.com/yandex-research/mind-your-format}}} 
    of prompt template sensitivity across 21 models and 4 datasets, showing that the performance gains similar to using in-context learning improvements can be achieved solely by selecting a proper template.
    \item We show that the choice of the best template depends on a combination of factors and that it is not possible to transfer the best template between models or prompting methods without a negative impact on quality.
    \item We propose Template Ensembles as a baseline solution for improving the template robustness for in-context learning.
\end{enumerate}

\section{Background and Related Work}
\label{sec:related_analysis}
\subsection{In-Context Learning}
\label{sec:icl}
An important property of LLMs is their ability to learn new tasks from only a few demonstrations~\cite{gpt2, gpt3}. 
This capability, known as in-context learning, forms the focus of our work.
We focus on sequence classification, as it is the most widely studied task for understanding and improving ICL performance.

Formally, classifying an input $x_{test}$ with in-context learning can be described as
finding the class $c$ in the space of label tokens $\mathcal{C}$ that yields a sequence with the highest probability according to a language model.
The input sequence consists of demonstration inputs and labels $(x_i, y_i)$ and a test input $(x_{test}, c)$; to obtain a natural language input, demonstrations are formatted with a template.

Each template consists of four components: input and output verbalizers $v_I(x)$ and $v_O(y, \mathcal{C})$ that transform $(x_i, y_i)$ into a natural language text, an intra-separator to divide input from output, and an inter-separator to join several demonstrations. 
\Cref{figure:template_transform} shows an example of transforming a set of demonstrations into a context for ICL.

\subsection{In-Context Learning Analysis}
Recent work has shown that ICL can perform at levels comparable to finetuning~\cite{chowdhery2022palm, chinchilla}.
Still, in-context learning is known to be highly dependent on the way the model input is formed: a prompt is defined by several components, and altering any of them can lead to unpredictable changes in performance.

\paragraph{Template Selection}
There are multiple ways to construct a template for a task. 
The most straightforward approach is to use minimal templates ($v_I = \{x\}, v_O = \{\mathcal{C}[y]\}$) or universal verbalizers like ``input/output'', as done in~\citet{implicitly_topic_models} and \citet{icl_differently}. 

Another strategy is to create task-specific templates. 
\citet{jiang-etal-2020-know} generate paraphrases of templates for the relation extraction task.
Authors show the sensitivity of masked language models to the prompt format and propose to ensemble predictions over the best templates.
Compared to this method, our approach is task-agnostic and does not require evaluating all templates in advance.

Several studies aim to find templates that directly optimize in-context learning performance~\cite{shin-etal-2020-autoprompt, gao-etal-2021-making}.
Our work unifies the results of previous research, using the verbalizers proposed by~\citet{gao-etal-2021-making}, as well as minimal and universal templates.

\paragraph{Choice and Order of Demonstrations}
The choice of examples for ICL is highly important, as they enable the model to condition on correct input and label distributions for the task~\cite{self_adaptive, influences,min-etal-2022-rethinking,chang-jia-2023-data}.
Furthermore, the order of examples also significantly affects the results and does not transfer between models even within the same family~\cite{order_matters,calibrate}.

In this work, we analyze two recent methods for selecting demonstrations.
\citet{implicitly_topic_models} propose learning latent concept variables for a task and using them to find examples that can best predict the task concept.
We refer to this method as \textit{Implicit Topic Models} or \textsc{ITM}.
In turn, \textsc{z-ICL}~\cite{z_icl} generates pseudo-demonstrations by retrieving most similar examples to the test sentence from an unlabeled dataset and assigning random labels to retrieved examples.

Crucially, both methods are evaluated on single templates that differ across two works.
Therefore, it is unclear whether the reported performance gains arise from the methods themselves or from a particular combination of the example selection strategy, the model, and the chosen template.

\paragraph{Prediction Methods}
\label{sec:related_prediction}
The standard approach for classification with LLMs is to compute the sequence probability with each of the possible labels and select the label with the highest probability.
We refer to this method as \textsc{Direct} further on.

Alternatively, one can use more advanced prediction methods that aim to reduce the variance across prompt formats.
The \textsc{Calibration} method~\cite{calibrate} computes a correction factor based on the deviation of the model's predictions for a placeholder input from a uniform distribution over labels and applies this factor to test set predictions.
Recent work has proposed multiple improvements of this method~\cite{fei-etal-2023-mitigating,han2023prototypical,zhou2024batch}; to limit the scope of our study, we focus only on the base \textsc{Calibration} approach in this work.
Lastly, the \textsc{Channel} prompting technique, proposed in~\citet{channel}, maximizes $P(x|y)$ instead of $P(y|x)$.

Both of these methods aim to mitigate the issue of ICL sensitivity to the prompt template choice.
However, as we show in~\Cref{app:prior_templates}, these methods are evaluated on their own sets of templates. 
In this paper, we strive for a more unified view on the robustness of advanced prompting methods and compare their performance across a broader range of templates and models.

\paragraph{Prompt and Template Robustness}
Although the problem of prompt robustness is relatively well-known, until recently, the discussion of \textit{template robustness} has been limited.
Notably, \citet{sclar2023quantifying} present a highly relevant study of prompt format sensitivity, reporting a significant performance variation across formats even for large models or minor template changes.
While their experiments are conducted in the standard setup (randomly selected examples and default prompting), our work instead focuses on alternative prompting and example selection methods, several of which~\cite{calibrate,channel} were proposed to improve the prompt robustness of ICL.
Similarly to papers in other subfields of machine learning arguing for a more consistent methodology~\cite{dacrema2019are,musgrave2020metric,platonov2023a}, the goal of our work is to demonstrate that disparate experiment setups lead to an invalid comparison of competing methods.

Moreover, several works study prompt robustness in a broader sense by considering models that use natural language instructions instead of labeled demonstrations~\cite{webson-pavlick-2022-prompt, leidinger-etal-2023-language,weber2023icl}.
Recently, \citet{mizrahi2023state} have shown that very similar instructions can lead to drastic differences in task performance for a variety of instruction-tuned models.
Although we study a similar issue, the focus of our work is on in-context learning and the transfer of best prompts between evaluation setups.
Still, we find that instruction-tuned models lack in-context robustness as well, which confirms previous observations and emphasizes the need for language model evaluation that takes prompt design into account.

\section{Setup \& Methodology}

\begin{table}[t]
    \centering
    \small
    \begin{tabular}{lc}
    \toprule
         Model family & Parameters (B) \\
         \midrule
         GPT-J~\cite{gptj} & 6 \\
         GPT-NeoX~\cite{gptneox} &  20 \\
         BLOOM~\cite{bloom} & 1.7, 3, 7.1 \\
         OPT~\cite{opt} & 6.7, 30, 66 \\
         Pythia~\cite{pythia} & 6.9, 12 \\
         LLaMA~\cite{llama} & 7, 13, 30, 65 \\
         Llama 2~\cite{llama2} & 13, 70 \\
         Falcon~\cite{falcon} & 1, 7, 40 \\
         \midrule
         Llama 3 Instruct~\cite{llama3} & 8 \\
         Mistral v0.3 Instruct~\cite{mistral} & 7 \\
         \bottomrule
    \end{tabular}
    \caption{Language models used in our work.}
    \label{table:models}
\vspace{-6pt}
\end{table}

\subsection{Models and Data}

We evaluate the robustness of in-context learning to template selection across a wide range of models on classification tasks.
All models used in our work are listed in~\Cref{table:models}: we run experiments on model families frequently used in literature (such as OPT and BLOOM), as well as the latest models with the highest quality (such as Llama 2 and Falcon).

In preliminary experiments, we observed that the performance of some models in the few-shot regime lags behind their zero-shot results. 
Hence, we excluded these models from further investigation. 
Further details regarding this selection procedure can be found in~\Cref{app:models_selection}.

We experiment with 4 sequence classification datasets: SST-2~\cite{socher-etal-2013-recursive}, DBPedia ontology classification task~\cite{dbpedia}, AGNews~\cite{agnews}, and TREC Question Classification~\cite{li-roth-2002-learning}. 
Although these datasets are frequently used in ICL studies, there is no consensus regarding the templates that should be used for each task.

\begin{table*}[hb!]
\setlength{\tabcolsep}{10pt}
    \centering
    \small
    \begin{tabular}{lcccc}
    \toprule
         Dataset & Input verbalizer & Output verbalizer & Intra-separator & Inter-separator \\
    \midrule
         \multirow{3}{*}{SST-2} &
         &
         ``output: \{\}'', ``target: \{\}'', ``label: \{\}'', &
         & 
         \\
         &
         \multirow{6}{*}{\Shortstack{``input: \{\}'', \\ ``text: \{\}'', \\ ``sentence: \{\}'', \\ ``\{\}''}} & 
         ``emotion: \{\}'', ``sentiment: \{\}'', ``A \{\} one.'' &
         \multirow{6}{*}{\Shortstack{`` '', \\ ``\textbackslash n''}} &
         \multirow{6}{*}{\Shortstack{`` '', \\ ``\textbackslash n'', \\ ``\textbackslash n\textbackslash n''}} \\
         &
         &
         ``It was \{\}.'', ``All in all \{\}.'', ``A \{\} piece.'' &
         &
         \\
         \cmidrule(lr){1-1}\cmidrule(lr){3-3}
         DBPedia &
         & 
         ``output: \{\}'', ``target: \{\}'', ``label: \{\}'', &
         & 
         \\
         \cmidrule(lr){1-1}
         AGNews &
         & 
         ``Topic: \{\}.'', ``Subject: \{\}.'',& 
         &
         \\
         \cmidrule(lr){1-1}
         TREC &
         &
         `This is about \{\}.'', ``It is about \{\}.''&
         &
         \\
     \bottomrule
    \end{tabular}
    \caption{Possible choices for all components of templates used in our work.}
    \label{table:templates}
\end{table*}

One can construct an input for in-context learning from a set of demonstrations by using a template consisting of four parts, as illustrated in~\Cref{figure:template_transform}. 
We present all options for verbalizers and separators for each dataset we study in~\Cref{table:templates}. 
Any combination of these components results in a valid template.
This set of options results in 216 possible prompt formats for SST-2 and 168 for DBPedia, AGNews and TREC.
A single evaluation run of all models on 10 random templates in one setup takes 17--48 hours on a single NVIDIA A100-80GB GPU, depending on the dataset.

\subsection{Methods}
Along with studying the robustness of standard in-context learning, we consider its improvements proposed in prior work. 
We focus on two main directions of ICL enhancements mentioned in~\Cref{sec:related_analysis}: example selection and prediction methods.
For each setting, we aggregate the results over 3 random seeds for example selection, with 10 random templates used for each seed and report the mean and standard deviation of classification accuracy, unless specified otherwise. 

As a baseline for demonstration selection, we choose the most straightforward approach of selecting $N$ random examples from the training dataset.
Intuitively, selecting more relevant examples for ICL should yield better performance. 
Therefore, we investigate the template sensitivity of two demonstration selection methods described in~\Cref{sec:icl}:~\textsc{ITM} and \textsc{z-ICL}.
Specifically, we select $N = 4$ examples using official implementations of each method.

Importantly, \textsc{ITM} requires training a concept model before choosing the examples. 
For GPT-2 Large, this procedure takes approximately 30 hours on a single NVIDIA A100-80GB. 
Repeating it for each model would be infeasible, especially given that the largest model has 86 times more parameters.
Therefore, we use the checkpoints of the GPT2 Large concept model provided by the authors to select demonstrations.
Also, we reuse the same examples for all models, leveraging authors' observations that demonstrations chosen with \textsc{ITM} can be transferred between models.

As discussed in~\Cref{sec:related_analysis}, more advanced prediction techniques can improve in-context learning accuracy. 
Therefore, we compare \textsc{Direct} prompting with \textsc{Channel}~\cite{channel} and \textsc{Calibration}~\cite{calibrate} prediction methods.

\section{Evaluation}
\label{sec:evaluation}

\subsection{Baseline Results}
We begin with analyzing the robustness of language models to the template choice in the baseline setup.
Specifically, we evaluate models in zero-shot and few-shot settings, selecting 2/4 random demonstrations and using the \textsc{Direct} prediction method.

Our results in~\Cref{table:baseline_main} show that even the most capable models such as Falcon and Llama 2 are highly sensitive to the prompt format;~\Cref{app:baseline_full} contains the results for the full set of 19 base models, and~\Cref{app:instruct_res} reports the results for instruction-tuned models.
Although the variance caused by this sensitivity makes it harder to observe the increase in ICL performance with the model size or the number of demonstrations, both trends still persist.
However, even the largest models have standard deviations of scores up to 35\% of their mean values.

To mitigate this lack of robustness, we could remove consistently underperforming prompt formats from the template pool.
We analyze the impact of separate template components in~\Cref{app:template-parts} and find that there are no specific parts (for example, verbalizers or separators) which could be excluded from evaluation.
Furthermore, we observe that a combination of ``suboptimal'' parts may result in an optimal template.

\begin{table*}[t]
    \centering
    \begin{tabular}{lcccccccccccc}
    \toprule
    \multirow{2}{*}{Model} &
    \multicolumn{2}{c}{SST-2} &
    \multicolumn{2}{c}{DBPedia} & 
    \multicolumn{2}{c}{AGNews} & 
    \multicolumn{2}{c}{TREC} \\
    
    \cmidrule(lr){2-3}\cmidrule(lr){4-5}\cmidrule(lr){6-7}
    \cmidrule(lr){8-9}
     & 2-shot & 4-shot & 2-shot & 4-shot & 2-shot & 4-shot & 2-shot & 4-shot
     \\
\midrule
Falcon 1B & 0.65\textsubscript{0.17} & 0.77\textsubscript{0.15} & 0.36\textsubscript{0.25} & 0.44\textsubscript{0.23} & 0.52\textsubscript{0.17} & 0.56\textsubscript{0.19} & 0.26\textsubscript{0.09} & 0.31\textsubscript{0.09} \\
Falcon 7B & 0.77\textsubscript{0.16} & 0.83\textsubscript{0.16} & 0.40\textsubscript{0.21} & 0.49\textsubscript{0.18} & 0.51\textsubscript{0.20} & 0.60\textsubscript{0.19} & 0.32\textsubscript{0.09} & 0.39\textsubscript{0.11} \\
Falcon 40B & 0.79\textsubscript{0.17} & 0.92\textsubscript{0.07} & 0.42\textsubscript{0.15} & 0.54\textsubscript{0.06} & 0.64\textsubscript{0.23} & 0.75\textsubscript{0.09} & 0.36\textsubscript{0.07} & 0.46\textsubscript{0.10} \\
\midrule
Llama 2 13B & 0.79\textsubscript{0.17} & 0.92\textsubscript{0.07} & 0.40\textsubscript{0.15} & 0.51\textsubscript{0.09} & 0.70\textsubscript{0.15} & 0.76\textsubscript{0.09} & 0.32\textsubscript{0.09} & 0.41\textsubscript{0.14} \\
Llama 2 70B & 0.83\textsubscript{0.14} & 0.92\textsubscript{0.09} & 0.46\textsubscript{0.15} & 0.60\textsubscript{0.05} & 0.76\textsubscript{0.14} & 0.82\textsubscript{0.05} & 0.41\textsubscript{0.07} & 0.51\textsubscript{0.06} \\

\bottomrule
    \end{tabular}
    \caption{Classification accuracy in the baseline setting for 2 LLM families. Standard deviation across 30 runs (10 templates for 3 sets of demonstrations) is in underscript. The results for all base models are presented in~\Cref{app:baseline_full}.}
    \label{table:baseline_main}
    \vspace{-12pt}
    \end{table*}
    
\subsection{Prediction Methods}
\label{sec:prediction_methods}
Next, we aim to evaluate the performance of different prediction methods in a unified setting.
Ideally, we would like these modifications to reduce the variance across templates, making the model behavior less dependent on the input format.

We evaluate \textsc{Channel} and \textsc{Calibration} methods in the 2-shot setting along with the \textsc{Direct} baseline. 
\begin{figure*}[]
    \centering
    \includegraphics[width=\linewidth]{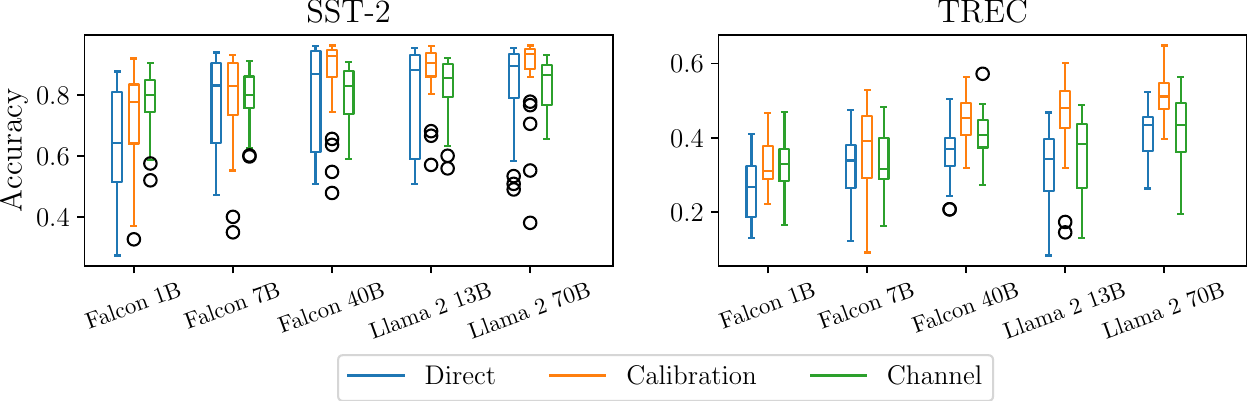}
    \caption{Comparison of in-context learning prediction methods in the 2-shot setting.}
    \label{fig:prediction_methods_main}
\vspace{-10pt}
\end{figure*}
As depicted in~\Cref{fig:prediction_methods_main}, both \textsc{Channel} and \textsc{Calibration} generally exhibit improved performance in comparison with \textsc{Direct}.
Still, for a number of models and datasets, the range of scores for \textsc{Direct} substantially overlaps with those of advanced methods.
This suggests that there are templates reaching the best performance with the \textsc{Direct} prediction method.

Additionally,~\Cref{table:prediction_methods_full} of~\Cref{app:prediction_methods_full} reveals that despite \textsc{Calibration} yielding the highest mean accuracy more often than other methods, it is more sensitive to the template choice than \textsc{Channel}.
Similar findings for instruction-tuned models are contained in~\Cref{app:instruct_res}.
Therefore, the choice of the prediction method should likely rely on the downstream usage scenario and the target evaluation setting.

\subsection{Example Selection Methods}
\label{sec:selection_methods}
Another area of ICL improvements that we evaluate on the matter of template sensitivity is the example selection strategy.
We compare \textsc{ITM} and \textsc{z-ICL} methods to the \textsc{Random} baseline in 4-shot setting, since using 4 demonstration was the main evaluation setting in the works proposing these methods. 
We use \textsc{Direct} prediction method to evaluate the gains of advanced example selection strategies independently from other ICL modifications.

\begin{figure*}[b]
    \centering
    \includegraphics[width=\linewidth]{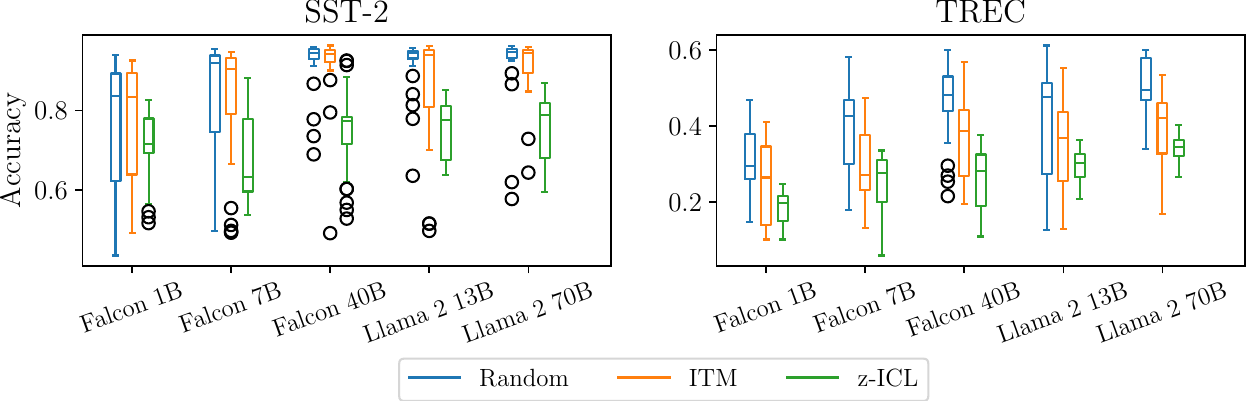}
    \caption{Comparison of the selection methods in the \textsc{Direct} 4-shot setting. For the evaluation results of other models and datasets, please refer to~\Cref{app:selection_methods_full}.}
    \label{fig:selection_methods_main}
\end{figure*}

Results in~\Cref{fig:selection_methods_main} and~\Cref{table:selection_methods_full} illustrate that when taking template sensitivity into account, advanced example selection methods often perform worse than random choice baseline.
\textsc{ITM} increases the average performance in most cases but still has a remarkably high standard deviation across templates.
Examples selected using the \textsc{z-ICL} method lead to more consistent but worse performance.

Note that our evaluation setup differs from those described in the original works, which might explain the discrepancy between our findings and the results reported by authors. 
Namely, we use the \textsc{Direct} prompting method and sample 10 random templates that may not include the templates used by authors of \textsc{ITM} and \textsc{z-ICL}.
To confirm template instability for prediction methods in their original implementations, we reproduce both methods and report our findings in~\Cref{app:reproduction}.
We observe high sensitivity to the prompt format, which raises a question of how much the reported gains of these methods can be attributed to the methods themselves and not to the template choice.

We conclude that the prompt format should be viewed as important as the example selection or the prediction method for ICL evaluation.
However, the search space of possible templates is infinite, which makes exhaustive search for each combination of the dataset, the model and the number of examples impractical.
Ideally, the best template for one setting would be optimal for all others or at least for similar settings. 
However, as we demonstrate in the following section, this is not the case.

\section{Template Transfer Evaluation}
\label{sec:transfer}
\subsection{Setup}
\label{sec:template_tuning}

We begin by defining a successful transfer between ICL settings.
In order to do so, we evaluate how the quality of model predictions varies across 30 random templates from~\Cref{table:templates}.
The results described in~\Cref{sec:templates-percent} demonstrate that the top-10 template on average yields 90\% of the best template score.
Therefore, if a prompt format is present in top-10 for both of the two compared setups, we can consider this an instance of successful transfer.

To compare sets of the best templates for a pair of settings, we compute Intersection-over-Union (IoU), also known as the Jaccard similarity coefficient ~\cite{jaccard12}, for top-10 best templates in each setting.
We also consider using the $\rho$ rank correlation coefficient~\cite{spearman04} as another measure of template transfer.
However, its value can increase when low-performing templates have similar rankings in different ICL setups, while the transfer of efficient templates remains low.
Still, we provide the results for this metric in~\Cref{app:transfer-spearman}.

\subsection{Transfer Between Models}
\begin{figure}
    \centering
    \includegraphics[width=\linewidth]{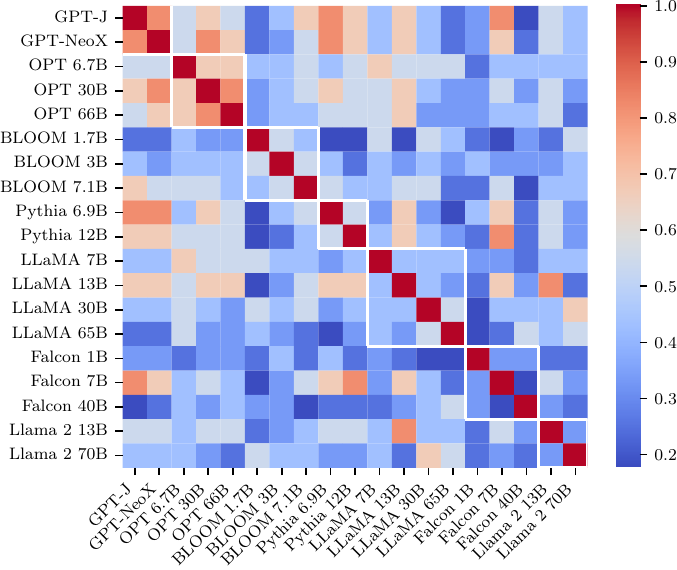}
    \caption{IoU of top-10 templates for all base models with 2 random demonstrations and the \textsc{Direct} prediction method on the DBPedia dataset.}
    \label{fig:transfer-models-iou-main}
    \vspace{-8pt}
\end{figure}

Next, we analyze the transfer of the best-performing templates between models in the baseline setup.
Specifically, we collect the results of each model in the 2-shot learning setting with \textsc{Direct} prediction method and \textsc{Random} demonstrations (fixed throughout the experiments) for 30 templates.
A heatmap of IoU for the transfer of top-10 best templates between \nmodels base models on the DBPedia dataset is presented in~\Cref{fig:transfer-models-iou-main};
for other datasets, please see~\Cref{sec:transfer-models}.

We observe that the IoU values exceed 0.5 only for a few model pairs on all datasets, meaning that the capacity for template transfer between models in the same setup is generally low.
This is especially concerning for models within a single family: as these models are trained on the same data and have the same architecture, one would expect them to perform similarly on the same prompt formats.

These observations signify that comparing ICL methods across models with a single template can lead to incorrect conclusions: a template that is effective for one model can easily be one of the worst choices for another model.

\subsection{Transfer Between Prediction Methods}
As discussed in~\Cref{sec:prediction_methods}, no prediction method that we evaluate can consistently outperform others across all models and datasets.
Therefore, to find an optimal setup for a new ICL improvement, one needs to evaluate every prediction technique in multiple templates.
We investigate the possibility of finding a universally optimal prompt for different methods to reduce the total computational cost.

 \begin{table}[t]
 \setlength{\tabcolsep}{3pt}
     \centering
     \begin{tabular}{lccc}
         \toprule
          \multirow{2}{*}{}         & Direct $\leftrightarrow$         & Direct $\leftrightarrow$        & Channel $\leftrightarrow$\\
                 & Calibration          & Channel        &  Calibration \\
         \midrule
         SST-2 & 0.49\textsubscript{0.17} & 0.30\textsubscript{0.11} & 0.31\textsubscript{0.08}\\
DBPedia & 0.54\textsubscript{0.17} & 0.47\textsubscript{0.15} & 0.45\textsubscript{0.14}\\
AG News & 0.36\textsubscript{0.11} & 0.25\textsubscript{0.13} & 0.35\textsubscript{0.14}\\
TREC & 0.31\textsubscript{0.12} & 0.23\textsubscript{0.09} & 0.28\textsubscript{0.13}\\
        
         \bottomrule
     \end{tabular}
     \caption{Intersection-over-Union for pairs of prompting methods averaged over the results of \nmodels base models obtained in the \textsc{Random} 2-shot setup.
     Standard deviations are in subscript.}
     \label{table:transfer-predmethods-iou}
     \vspace{-6pt}
 \end{table}

To answer this question, we calculate the IoU between top 10 performing templates for each method for a fixed set of demonstrations. 
Results in~\Cref{table:transfer-predmethods-iou} display that similarly to the models, the transfer between prediction methods is also low. 
Consequently, the prompt format sensitivity issue creates a burden on authors of new ICL modifications; they must tune templates for every prediction method they want to combine with their own approach.

\subsection{Transfer Between Demonstration Selection Methods}
Having found that the best-performing templates are specific both to the model and the prediction method,
we now aim to find whether the best formats would be the same for different demonstration sets in the same setup.
Similarly to previous experiments, we calculate IoU for 10 templates that yield the highest scores for each method.

\begin{figure}[h]
    \centering
    \includegraphics[width=\linewidth
    ]{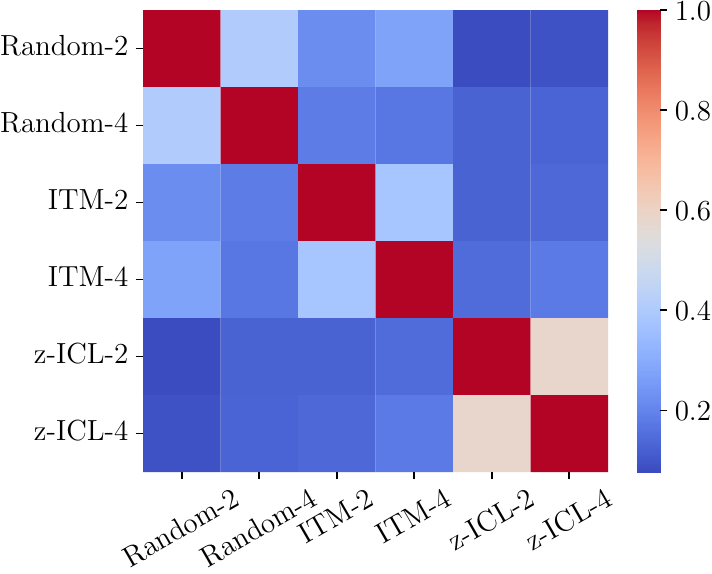}
    \caption{IoU of 10 best templates for example selection methods on the AG News dataset. \textsc{Method-N} indicates that \textsc{Method} was used to select $N$ examples.}
    \label{fig:selection_methods_iou}
    \vspace{-6pt}
\end{figure}

Results in~\Cref{fig:selection_methods_iou} illustrate that simply adding demonstrations, even if they were obtained with the same method, can significantly alter the ranking of the best templates.
This justifies the necessity to evaluate example selection methods on a range of templates to avoid misinterpretation of the results.

\subsection{Discussion}
Based on the above findings, we conclude that the results of evaluation of various ICL improvements without consideration of template sensitivity issue are hardly reliable for several reasons. 
First, as the best templates do not transfer between models even within the same family, scoring a method across several models using the same format will inevitably lead to underestimation of the method for all models except the one for which the format was tuned.
Next, as there is little evidence of transfer between setups, the format selection procedure needs to be precisely described and applied in all evaluated settings for a fair comparison.

In summary, we find that there are no universally well-performing prompt formats.
Therefore, the results of in-context learning evaluation can be reliable only if they are \textbf{aggregated over several templates} or if \textbf{each setting is evaluated in its best-performing template}.
The former approach requires accounting for the variance of the scores and makes comparison less apparent, while the latter can be computationally expensive.

\section{Template Ensembles}
\label{sec:ensemble}

\begin{table*}[hb!]
\vspace{-2pt}
\centering
\begin{tabular}{lcccccc}
\toprule
\multirow{2}{*}{Model} & \multicolumn{2}{c}{Direct} & \multicolumn{2}{c}{Channel} & \multicolumn{2}{c}{Calibration} \\
           \cmidrule(lr){2-3}\cmidrule(lr){4-5}\cmidrule(lr){6-7}
     &  Single &      Ensemble &  Single &  Ensemble & Single &   Ensemble \\
\midrule
LLaMA 2 13B &   0.79\textsubscript{0.17}  & 0.85\textsubscript{0.09} &   0.82\textsubscript{0.10}  & 0.90\textsubscript{0.02} &    0.88\textsubscript{0.09}  & \textbf{0.93}\textsubscript{0.03} \\
LLaMA 2 70B &   0.83\textsubscript{0.14}  & \textbf{0.95}\textsubscript{0.01} &   0.83\textsubscript{0.08}  & 0.92\textsubscript{0.01} &    0.88\textsubscript{0.13}  & 0.94\textsubscript{0.03} \\
\midrule
Falcon 1B &   0.65\textsubscript{0.17}  & 0.74\textsubscript{0.07} &   0.77\textsubscript{0.10}  & \textbf{0.89}\textsubscript{0.01} &    0.71\textsubscript{0.17}  & \textbf{0.89}\textsubscript{0.01} \\
Falcon 7B &   0.77\textsubscript{0.16}  & 0.81\textsubscript{0.00} &   0.78\textsubscript{0.09}  & 0.90\textsubscript{0.00} &    0.79\textsubscript{0.15}  & \textbf{0.93}\textsubscript{0.02} \\
Falcon 40B &   0.79\textsubscript{0.17}  & 0.93\textsubscript{0.01} &   0.81\textsubscript{0.09}  & 0.91\textsubscript{0.01} &    0.87\textsubscript{0.13}  & \textbf{0.95}\textsubscript{0.00} \\
\bottomrule
\end{tabular}
\caption{Comparison of 2-shot learning performance on the SST-2 dataset using ensembles of 5 templates and a single template. Standard deviations over 5 random seeds are in subscript, best accuracy for each model is in bold.}
\label{table:tmpl-ensembles}
\end{table*}

To reduce the variance in performance caused by the template choice, we propose to ensemble model predictions across multiple templates.
This approach is widely used in machine learning~\cite{ho1995random, lakshminarayanan2017simple} for improving the predictive performance of the model, as well as its robustness, and can be viewed as a form of test-time augmentation~\cite{krizhevsky2012imagenet,simonyan2015deep}.
Prior work on prompt ensembling has shown significant gains by training a boosting algorithm on model outputs~\cite{pmlr-v202-hou23b}; by contrast, our method needs only the pretrained model predictions without additional training.

Formally, our method computes label probabilities across predictions for each of $N$ templates, where $N$ is the ensemble size, and outputs the label with the highest average probability. 
In early experiments, we tried selecting the most common label among the predictions; however, we found this voting strategy to perform poorly on tasks with a large number of classes.
It is also important to note that ensembling $N$ predictions involves running the model $N$ times more compared to single-format evaluation, which makes this approach more computationally expensive.
We view template ensembles as a baseline solution for the problem of prompt format sensitivity and leave the exploration of more efficient methods to future work.

\begin{figure}[t]
    \centering
    \includegraphics[width=\linewidth]{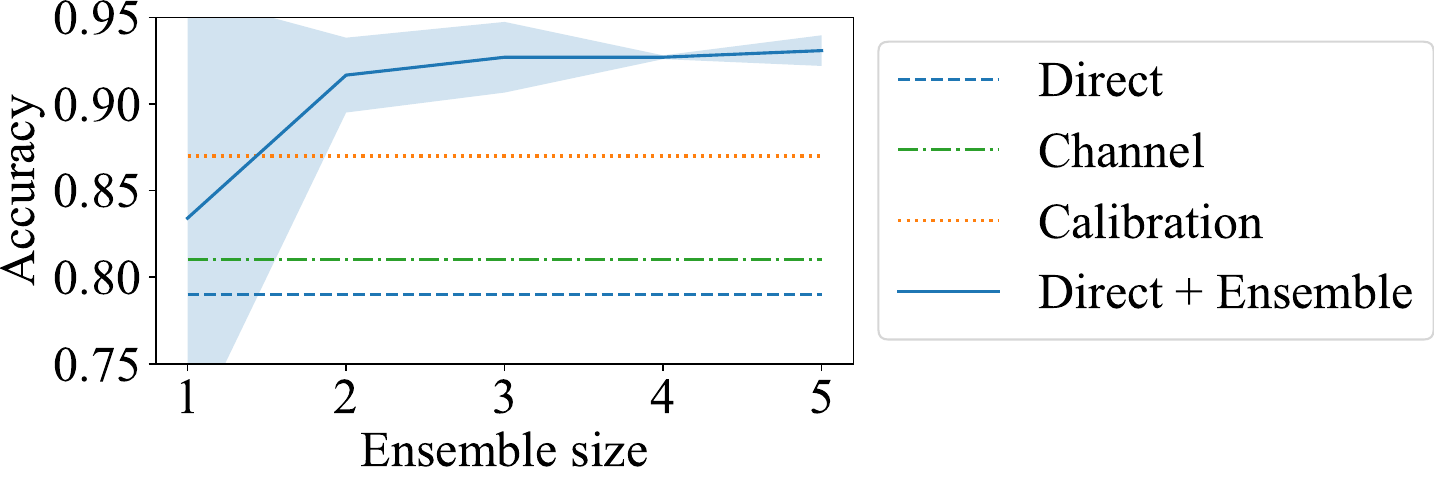}
    \caption{Template ensemble accuracy as a function of its size for Falcon 40B on the SST-2 dataset in the 2-shot learning setup. Dashed lines depict the results of baseline methods averaged over 10 templates.}
    \label{fig:falcon_ensemble}
    \vspace{-6pt}
\end{figure}

We begin with determining the minimal ensemble size that consistently reduces variance while increasing the average performance.
We observe that for the majority of models and prediction methods, an ensemble achieves the best accuracy when its size reaches 4 or 5 (see an example in~\Cref{fig:falcon_ensemble}), with further expansion being less effective.
We also found that smaller ensembles may demonstrate unstable behavior, with the possibility of a drop in performance if a suboptimal template is sampled. 
Therefore, we report results for ensembles of size 5 and average the results over 5 random seeds.

Next, we evaluate the performance gains of Template Ensembles for different prediction methods.
Our findings in~\Cref{table:tmpl-ensembles,table:ensembles_results_instruct} and~\Cref{app:ensembles-full,app:instruct_res} indicate that ensembles increase the accuracy for all evaluated models and prediction methods.
Most importantly, they also significantly reduce the variance caused by the template choice for most setups. 
Therefore, we conclude that template ensembling allows to preserve the increase in accuracy provided by ICL modifications while mitigating the template sensitivity issue.

\section{Conclusion}

In this work, we study the inconsistencies in the evaluation of in-context learning advancements introduced by the template sensitivity of large language models.
Specifically, we find that ICL improvements exhibit high variation across template formats and that it is not possible to reuse the same template across different modifications.
This aspect is often overlooked in prior work, despite the fact that the impact of template selection on prediction accuracy may be comparable with the choice of demonstrations or prompting methods.

While we propose Template Ensembles as an initial solution to this problem, the general sensitivity of language models to minor prompt variations is yet to be addressed.
Consequently, we believe that the research community should take this problem into account when developing new models, evaluation benchmarks, or in-context learning methods.

\section*{Limitations}
Due to limited computational resources and the high cost for evaluation on a large range of models, we only focus on four classification datasets. 
Moreover, we only compare two example selection methods to a random baseline, potentially overlooking other effective approaches. 

Additionally, the space of templates could be expanded for more comprehensive experimentation. 
For example, we did not explore label mapping, including random labels, which is an important aspect of the template.

We would like to notice that our study focuses on a template selection impact on a performance and a degree of template transfer between different setups but not on templates themselves. 
Future work should further analyze not only which templates lead to a change in performance but also on why they affect it.

\bibliography{anthology,custom}

\appendix
\clearpage
\section{Templates from Prior Work}
\label{app:prior_templates}
\Cref{table:all_templates_sst2,table:all_templates_dbpedia,table:all_templates_trec,table:all_templates_agnews} provide a comparison of all the templates used in the works presenting all methods we evaluate.
Noticeably, prompt formats (and the choice of label words for some formats and datasets) used in works proposing investigated methods have no intersection.
This is also concerning, since the original papers proposing these methods refer to each other.
For instance, \textsc{Channel} prompting outperforming \textsc{Calibration} in~\citet{channel} might be explained by selecting a more favorable set of templates for the method proposed in the paper rather than by the advantages of the method itself.

\begin{table*}[]
\setlength{\tabcolsep}{5pt}
\centering
\begin{tabular}{lccccc}
\toprule
Method & Input verbalizer & Output verbalizer &  Intra-sep & Inter-sep & Label words \\
\midrule 
ITM & ``sentence: \{\}'' & ``\{\}'' & `` " & `` '' & negative, positive \\
\midrule
z-ICL & ``Review: \{\}'' & ``Sentiment: \{\}'' & ``\textbackslash n'' & ``\textbackslash n\textbackslash n\textbackslash n'' & terrible, great \\
\midrule
\multirow{4}{*}{Channel} & ``\{\}'' & ``A \{\} one'' & `` "& `` ''& terrible, great \\
&  ``\{\}'' & ``It was \{\}.'' & `` ''& `` ''& terrible, great\\
&  ``\{\}'' & ``All in all \{\}.'' & `` ''& `` ''& terrible, great\\
&  ``\{\}'' & ``A \{\} piece.'' & `` ''& `` ''& terrible, great\\
\midrule
\multirow{30}{*}{Calibration} & ``Review: \{\}'' & ``Answer: \{\}'' & ``\textbackslash n'' & ``\textbackslash n\textbackslash n'' & Negative, Positive \\
& ``Review: \{\}'' & ``Answer: \{\}'' & ``\textbackslash n'' & ``\textbackslash n\textbackslash n'' & bad, good \\
& ``Review: \{\}'' & ``Positive review? \{\}'' & ``\textbackslash n'' & ``\textbackslash n\textbackslash n'' & No, Yes \\
& ``Input: \{\}'' & ``Sentiment: \{\}'' & ``\textbackslash n'' & ``\textbackslash n\textbackslash n'' & Negative, Positive \\
& ``Review: \{\}'' & ``Positive: \{\}'' & ``\textbackslash n'' & ``\textbackslash n\textbackslash n'' & False, True \\
& \makecell{``My review for last \\night’s film: \{\}''} & \makecell{``The critics agreed that \\ this movie was \{\}''} & `` '' & ``\textbackslash n\textbackslash n'' & bad, good \\
& \makecell{``One of our critics \\wrote \{\}''} & \makecell{``Her sentiment towards \\ the film was \{\}''} & `` '' & ``\textbackslash n\textbackslash n'' & Negative, Positive \\
& \makecell{``In a contemporary \\review, Roger Ebert \\ wrote \{\}.''} & \makecell{``Entertainment Weekly \\agreed, and 
the overall \\critical reception of \\ the film was \{\}''} & `` '' & ``\textbackslash n\textbackslash n'' & bad, good \\
& ``Review: \{\}'' & \makecell{``Question: Is the \\sentiment of the \\above review\\Positive or Negative?\\\textbackslash nAnswer: \{\}''} & ``\textbackslash n'' & ``\textbackslash n\textbackslash n'' & Negative, Positive\\
& ``Review: \{\}'' & \makecell{``Question: Did the\\author think that the \\movie was good or\\ bad?\textbackslash nAnswer: \{\}''} & ``\textbackslash n'' & ``\textbackslash n\textbackslash n'' & bad, good\\
& \makecell{``Question: Did the \\author of the\\ following tweet \\think that the \\ movie was good \\or bad?\textbackslash nTweet: \{\}''} & ``Answer: \{\}'' & ``\textbackslash n'' & ``\textbackslash n\textbackslash n'' & bad, good \\
& ``\{\}'' & \makecell{``My overall feeling \\was that the \\movie was \{\}''} & `` '' & ``\textbackslash n\textbackslash n'' & bad, good \\
& ``\{\}'' & ``I \{\} the movie.''& `` '' & ``\textbackslash n\textbackslash n'' & hated, liked \\
& ``\{\}'' & \makecell{``My friend asked \\me if I would \\give the movie 0 or \\5 stars, I said \{\}''} & `` '' & ``\textbackslash n\textbackslash n'' & 0, 5 \\
\bottomrule
\end{tabular}
\caption{All templates used in methods we evaluate for SST-2 dataset.}
\label{table:all_templates_sst2}
\end{table*}

\begin{table*}[]
\centering
\setlength{\tabcolsep}{5pt}
\begin{tabular}{lccccc}
\toprule
Method & Input verbalizer & Output verbalizer &  Intra-sep & Inter-sep & Label words \\
\midrule 
Channel & 
\makecell{``\{\}''\\``\{\}''\\``\{\}''\\``\{\}''} & 
\makecell{``Topic:  \{\}.''\\ ``Subject: \{\}.''\\ ``This is about \{\}.''\\``It is about \{\} one.''} &
\makecell{`` ''\\`` ''\\`` ''\\`` ''} & 
\makecell{`` ''\\`` ''\\`` ''\\`` ''} & 
\makecell{
     Company,\\ Educational Institution,\\ Artist, Athlete,\\ Office Holder,\\ Building,\\ Natural Place, Village,\\ Animal, Plant,\\ Album, Film,\\ Written Work,\\ Mean of Transportation} \\
\midrule
ITM & ``\{\}'' & ``\{\}'' & `` '' & `` '' & Same as above \\
\midrule
Calibration & \makecell{``Classify the \\documents based \\ on whether they\\ are about \\ a [Label words]\\ \textbackslash n\textbackslash n Article: \{\}''} 
& ``Answer: \{\}''& ``\textbackslash n'' & ``\textbackslash n\textbackslash n'' & \makecell{Company, School,\\ Artist, Athlete,\\ Politician,\\ Building, Nature,\\ Village, Animal,\\ Plant, Album,\\ Film, Book,\\ Transportation} \\
\bottomrule
\end{tabular}
\caption{All templates used in methods we evaluate for DBPedia dataset.}
\label{table:all_templates_dbpedia}
\end{table*}

\begin{table*}[]
\setlength{\tabcolsep}{5pt}
\centering
\begin{tabular}{lccccc}
\toprule
Method & Input verbalizer & Output verbalizer &  Intra-sep & Inter-sep & Label words \\
\midrule 
\multirow{4}{*}{Channel} & ``\{\}'' & ``\{\}'' & `` '' & `` '' & 
\multirow{4}{*}{
\begin{tabular}{@{}c@{}}
Description, Entity,\\ Expression, Human,\\ Location, Number
\end{tabular}}\\
&  ``\{\}'' & ``Q: \{\}." & `` ''& `` ''& \\
&  ``\{\}'' & ``Why \{\}?" & `` ''& `` ''& \\
&  ``\{\}'' & ``Answer: \{\}'' & `` ''& `` ''& \\
\midrule
Calibration & \makecell{``Classify the \\questions  based \\ on whether their\\ answer type is \\a [Label words]\textbackslash n\textbackslash n\\
Question: \{\}''} & ``Answer Type: \{\}'' & ``\textbackslash n'' & ``\textbackslash n\textbackslash n'' & \makecell{Number, Location,\\ Person, Description,\\Entity, Abbreviation} \\
\bottomrule
\end{tabular}
\caption{All templates used in methods we evaluate for TREC dataset.}
\label{table:all_templates_trec}
\end{table*}

\begin{table*}[]
\centering
\setlength{\tabcolsep}{5pt}
\begin{tabular}{lccccc}
\toprule
Method & Input verbalizer & Output verbalizer &  Intra-sep & Inter-sep & Label words \\
\midrule 
\multirow{4}{*}{Channel} & ``\{\}'' & ``Topic:  \{\}." & `` '' & `` '' & \multirow{4}{*}{
\begin{tabular}{@{}c@{}}
World, Sports,\\ Business, Technology
\end{tabular}}\\
&  ``\{\}'' & ``Subject: \{\}." & `` ''& `` ''& \\
&  ``\{\}'' & ``This is about \{\}." & `` ''& `` ''& \\
&  ``\{\}'' & ``It is about \{\} one." & `` ''& `` ''& \\
\midrule
Calibration & ``Article: \{\}'' & ``Answer: \{\}'' & ``\textbackslash n'' & ``\textbackslash n\textbackslash n'' & Same as above. \\
\bottomrule
\end{tabular}
\caption{All templates used in methods we evaluate for AG News dataset.}
\label{table:all_templates_agnews}
\end{table*}

\section{Model Selection}
\label{app:models_selection}
Our initial evaluation pool consisted of 23 models. 
We evaluated each of them in 0-shot and 2-shot settings with three prediction methods on four datasets, resulting in 12 runs. 
For each run in both 0-shot and 2-shot setups, we compare the model performance averaged over 10 random templates.

Based on the results presented in~\Cref{table:prediction_methods_full}, we restricted the final pool of models for evaluation to those that have a consistent increase in performance in the 2-shot setting, in other words, to those demonstrating a performance boost from ICL. 
More specifically, we kept the models that had 8 or more wins in 2-shot evaluation against 0-shot.

\begin{table*}[]
    \centering
    \small
    \setlength{\tabcolsep}{1.55pt}
    \begin{tabular}{lcccccccccccccc}
    \toprule
    \multirow{2}{*}{Model} & \multirow{2}{*}{N} & \multicolumn{3}{c}{SST-2} & \multicolumn{3}{c}{DBPedia} & \multicolumn{3}{c}{AGNews} & \multicolumn{3}{c}{TREC} & \multirow{2}{*}{\makecell{2-shot\\ wins}} \\
    \cmidrule(lr){3-5}\cmidrule(lr){6-8}\cmidrule(lr){9-11}\cmidrule(lr){12-14}
     & & Direct & Channel & Calib. & Direct & Channel & Calib. & Direct & Channel & Calib. & Direct & Channel & Calib. \\
\midrule
\rowcolor{Gray}
& 0 & 0.65\textsubscript{0.09} & 0.72\textsubscript{0.04} & 0.70\textsubscript{0.06} & 0.34\textsubscript{0.13} & 0.40\textsubscript{0.07} & 0.49\textsubscript{0.09} & 0.48\textsubscript{0.16} & 0.56\textsubscript{0.04} & 0.64\textsubscript{0.14} & 0.25\textsubscript{0.06} & 0.28\textsubscript{0.11} & 0.30\textsubscript{0.08} & \\
\rowcolor{Gray}
\multirow{-2}{*}{GPT-2 Large} & 2 & 0.59\textsubscript{0.10} & 0.70\textsubscript{0.12} & 0.62\textsubscript{0.08} & 0.14\textsubscript{0.08} & 0.53\textsubscript{0.10} & 0.58\textsubscript{0.10} & 0.32\textsubscript{0.12} & 0.58\textsubscript{0.07} & 0.57\textsubscript{0.09} & 0.26\textsubscript{0.09} & 0.29\textsubscript{0.08} & 0.30\textsubscript{0.06} & \multirow{-2}{*}{6/12} \\
\rowcolor{Gray}
& 0 & 0.76\textsubscript{0.04} & 0.73\textsubscript{0.05} & 0.70\textsubscript{0.09} & 0.40\textsubscript{0.05} & 0.43\textsubscript{0.08} & 0.54\textsubscript{0.09} & 0.52\textsubscript{0.09} & 0.56\textsubscript{0.05} & 0.65\textsubscript{0.08} & 0.23\textsubscript{0.03} & 0.24\textsubscript{0.07} & 0.26\textsubscript{0.05} &  \\
\rowcolor{Gray}
\multirow{-2}{*}{GPT-2 XL} & 2 & 0.58\textsubscript{0.11} & 0.71\textsubscript{0.09} & 0.63\textsubscript{0.11} & 0.15\textsubscript{0.09} & 0.54\textsubscript{0.09} & 0.50\textsubscript{0.15} & 0.40\textsubscript{0.20} & 0.61\textsubscript{0.08} & 0.56\textsubscript{0.15} & 0.26\textsubscript{0.07} & 0.34\textsubscript{0.09} & 0.33\textsubscript{0.08} & \multirow{-2}{*}{5/12}\\
\midrule
\multirow{2}{*}{GPT-J} & 0 & 0.71\textsubscript{0.09} & 0.68\textsubscript{0.08} & 0.68\textsubscript{0.08} & 0.41\textsubscript{0.07} & 0.44\textsubscript{0.06} & 0.57\textsubscript{0.08} & 0.61\textsubscript{0.08} & 0.64\textsubscript{0.03} & 0.64\textsubscript{0.07} & 0.32\textsubscript{0.05} & 0.20\textsubscript{0.07} & 0.33\textsubscript{0.04} & \multirow{2}{*}{9/12} \\
& 2 & 0.65\textsubscript{0.14} & 0.77\textsubscript{0.11} & 0.68\textsubscript{0.11} & 0.25\textsubscript{0.16} & 0.68\textsubscript{0.06} & 0.71\textsubscript{0.16} & 0.47\textsubscript{0.19} & 0.67\textsubscript{0.09} & 0.73\textsubscript{0.11} & 0.26\textsubscript{0.07} & 0.32\textsubscript{0.09} & 0.33\textsubscript{0.06} \\
\multirow{2}{*}{GPT-NeoX} & 0 & 0.71\textsubscript{0.08} & 0.67\textsubscript{0.06} & 0.70\textsubscript{0.09} & 0.48\textsubscript{0.04} & 0.42\textsubscript{0.05} & 0.60\textsubscript{0.07} & 0.67\textsubscript{0.06} & 0.56\textsubscript{0.04} & 0.58\textsubscript{0.07} & 0.30\textsubscript{0.08} & 0.22\textsubscript{0.06} & 0.32\textsubscript{0.05} & \multirow{2}{*}{9/12}\\
& 2 & 0.69\textsubscript{0.15} & 0.82\textsubscript{0.06} & 0.79\textsubscript{0.12} & 0.32\textsubscript{0.19} & 0.67\textsubscript{0.05} & 0.72\textsubscript{0.16} & 0.52\textsubscript{0.22} & 0.67\textsubscript{0.10} & 0.70\textsubscript{0.13} & 0.31\textsubscript{0.08} & 0.32\textsubscript{0.07} & 0.36\textsubscript{0.08} \\
\midrule
\rowcolor{Gray}
 & 0 & 0.78\textsubscript{0.07} & 0.68\textsubscript{0.07} & 0.79\textsubscript{0.07} & 0.41\textsubscript{0.05} & 0.33\textsubscript{0.08} & 0.57\textsubscript{0.12} & 0.49\textsubscript{0.07} & 0.60\textsubscript{0.01} & 0.67\textsubscript{0.07} & 0.27\textsubscript{0.04} & 0.17\textsubscript{0.06} & 0.24\textsubscript{0.03} &  \\
\rowcolor{Gray}
\multirow{-2}{*}{OPT 1.3B} & 2 & 0.69\textsubscript{0.15} & 0.80\textsubscript{0.06} & 0.71\textsubscript{0.16} & 0.21\textsubscript{0.11} & 0.58\textsubscript{0.08} & 0.61\textsubscript{0.12} & 0.48\textsubscript{0.24} & 0.66\textsubscript{0.06} & 0.61\textsubscript{0.11} & 0.27\textsubscript{0.08} & 0.38\textsubscript{0.09} & 0.35\textsubscript{0.08} & \multirow{-2}{*}{6/12} \\
\multirow{2}{*}{OPT 6.7B} & 0 & 0.79\textsubscript{0.07} & 0.67\textsubscript{0.07} & 0.80\textsubscript{0.07} & 0.46\textsubscript{0.04} & 0.49\textsubscript{0.05} & 0.61\textsubscript{0.06} & 0.59\textsubscript{0.08} & 0.61\textsubscript{0.06} & 0.64\textsubscript{0.07} & 0.24\textsubscript{0.04} & 0.27\textsubscript{0.09} & 0.33\textsubscript{0.02} & \multirow{2}{*}{9/12}\\
& 2 & 0.67\textsubscript{0.16} & 0.81\textsubscript{0.06} & 0.72\textsubscript{0.19} & 0.27\textsubscript{0.14} & 0.69\textsubscript{0.05} & 0.71\textsubscript{0.17} & 0.45\textsubscript{0.17} & 0.69\textsubscript{0.09} & 0.70\textsubscript{0.14} & 0.27\textsubscript{0.08} & 0.34\textsubscript{0.08} & 0.34\textsubscript{0.07} \\
\multirow{2}{*}{OPT 30B} & 0 & 0.79\textsubscript{0.06} & 0.72\textsubscript{0.05} & 0.77\textsubscript{0.08} & 0.48\textsubscript{0.04} & 0.48\textsubscript{0.07} & 0.61\textsubscript{0.08} & 0.64\textsubscript{0.05} & 0.60\textsubscript{0.06} & 0.65\textsubscript{0.11} & 0.24\textsubscript{0.03} & 0.26\textsubscript{0.05} & 0.31\textsubscript{0.01} & \multirow{2}{*}{8/12}\\
& 2 & 0.64\textsubscript{0.17} & 0.79\textsubscript{0.09} & 0.73\textsubscript{0.17} & 0.34\textsubscript{0.21} & 0.73\textsubscript{0.05} & 0.78\textsubscript{0.14} & 0.55\textsubscript{0.19} & 0.69\textsubscript{0.09} & 0.76\textsubscript{0.11} & 0.31\textsubscript{0.06} & 0.35\textsubscript{0.09} & 0.33\textsubscript{0.06} \\
\multirow{2}{*}{OPT 66B} & 0 & 0.73\textsubscript{0.12} & 0.73\textsubscript{0.07} & 0.74\textsubscript{0.10} & 0.41\textsubscript{0.03} & 0.48\textsubscript{0.07} & 0.61\textsubscript{0.09} & 0.64\textsubscript{0.07} & 0.58\textsubscript{0.06} & 0.62\textsubscript{0.07} & 0.26\textsubscript{0.03} & 0.23\textsubscript{0.06} & 0.31\textsubscript{0.05} & \multirow{2}{*}{8/12}\\
& 2 & 0.65\textsubscript{0.15} & 0.81\textsubscript{0.08} & 0.76\textsubscript{0.16} & 0.34\textsubscript{0.16} & 0.77\textsubscript{0.06} & 0.81\textsubscript{0.15} & 0.45\textsubscript{0.17} & 0.74\textsubscript{0.05} & 0.70\textsubscript{0.14} & 0.28\textsubscript{0.07} & 0.38\textsubscript{0.08} & 0.34\textsubscript{0.07} \\
\midrule
\multirow{2}{*}{BLOOM 1.7B} & 0 & 0.68\textsubscript{0.11} & 0.67\textsubscript{0.06} & 0.68\textsubscript{0.11} & 0.47\textsubscript{0.03} & 0.47\textsubscript{0.06} & 0.47\textsubscript{0.07} & 0.61\textsubscript{0.08} & 0.53\textsubscript{0.04} & 0.58\textsubscript{0.06} & 0.27\textsubscript{0.04} & 0.24\textsubscript{0.08} & 0.33\textsubscript{0.03} & \multirow{2}{*}{9/12}\\
& 2 & 0.66\textsubscript{0.12} & 0.75\textsubscript{0.06} & 0.71\textsubscript{0.10} & 0.27\textsubscript{0.19} & 0.62\textsubscript{0.08} & 0.57\textsubscript{0.13} & 0.43\textsubscript{0.19} & 0.59\textsubscript{0.08} & 0.61\textsubscript{0.11} & 0.31\textsubscript{0.09} & 0.39\textsubscript{0.07} & 0.37\textsubscript{0.08} \\
\multirow{2}{*}{BLOOM 3B} & 0 & 0.71\textsubscript{0.10} & 0.71\textsubscript{0.06} & 0.70\textsubscript{0.08} & 0.39\textsubscript{0.06} & 0.40\textsubscript{0.07} & 0.48\textsubscript{0.05} & 0.66\textsubscript{0.02} & 0.48\textsubscript{0.06} & 0.60\textsubscript{0.08} & 0.22\textsubscript{0.06} & 0.20\textsubscript{0.07} & 0.20\textsubscript{0.06} & \multirow{2}{*}{9/12}\\
& 2 & 0.72\textsubscript{0.14} & 0.77\textsubscript{0.09} & 0.77\textsubscript{0.10} & 0.27\textsubscript{0.21} & 0.67\textsubscript{0.06} & 0.57\textsubscript{0.14} & 0.45\textsubscript{0.19} & 0.62\textsubscript{0.07} & 0.67\textsubscript{0.13} & 0.34\textsubscript{0.09} & 0.35\textsubscript{0.08} & 0.36\textsubscript{0.09} \\
\multirow{2}{*}{BLOOM 7.1B} & 0 & 0.72\textsubscript{0.09} & 0.71\textsubscript{0.06} & 0.68\textsubscript{0.06} & 0.44\textsubscript{0.05} & 0.45\textsubscript{0.08} & 0.51\textsubscript{0.08} & 0.64\textsubscript{0.06} & 0.56\textsubscript{0.04} & 0.64\textsubscript{0.10} & 0.35\textsubscript{0.07} & 0.22\textsubscript{0.08} & 0.32\textsubscript{0.04} & \multirow{2}{*}{9/12}\\
& 2 & 0.69\textsubscript{0.15} & 0.76\textsubscript{0.09} & 0.76\textsubscript{0.11} & 0.26\textsubscript{0.18} & 0.70\textsubscript{0.06} & 0.67\textsubscript{0.14} & 0.43\textsubscript{0.17} & 0.69\textsubscript{0.06} & 0.68\textsubscript{0.12} & 0.33\textsubscript{0.08} & 0.34\textsubscript{0.07} & 0.36\textsubscript{0.06} \\
\midrule
\multirow{2}{*}{Pythia 6.9B} & 0 & 0.75\textsubscript{0.08} & 0.72\textsubscript{0.05} & 0.69\textsubscript{0.11} & 0.45\textsubscript{0.05} & 0.43\textsubscript{0.04} & 0.63\textsubscript{0.09} & 0.58\textsubscript{0.14} & 0.59\textsubscript{0.04} & 0.64\textsubscript{0.08} & 0.31\textsubscript{0.07} & 0.21\textsubscript{0.07} & 0.32\textsubscript{0.03} & \multirow{2}{*}{8/12}\\
& 2 & 0.63\textsubscript{0.12} & 0.78\textsubscript{0.09} & 0.77\textsubscript{0.11} & 0.28\textsubscript{0.16} & 0.67\textsubscript{0.08} & 0.68\textsubscript{0.14} & 0.43\textsubscript{0.17} & 0.68\textsubscript{0.09} & 0.69\textsubscript{0.14} & 0.34\textsubscript{0.09} & 0.37\textsubscript{0.07} & 0.38\textsubscript{0.06} \\
\multirow{2}{*}{Pythia 12B} & 0 & 0.73\textsubscript{0.07} & 0.71\textsubscript{0.08} & 0.69\textsubscript{0.10} & 0.43\textsubscript{0.05} & 0.43\textsubscript{0.04} & 0.51\textsubscript{0.18} & 0.61\textsubscript{0.09} & 0.57\textsubscript{0.05} & 0.65\textsubscript{0.09} & 0.33\textsubscript{0.06} & 0.23\textsubscript{0.05} & 0.32\textsubscript{0.03} & \multirow{2}{*}{8/12}\\
& 2 & 0.63\textsubscript{0.13} & 0.79\textsubscript{0.10} & 0.74\textsubscript{0.12} & 0.29\textsubscript{0.15} & 0.68\textsubscript{0.07} & 0.71\textsubscript{0.14} & 0.53\textsubscript{0.18} & 0.68\textsubscript{0.08} & 0.70\textsubscript{0.12} & 0.29\textsubscript{0.09} & 0.35\textsubscript{0.06} & 0.33\textsubscript{0.08} \\
\midrule
\multirow{2}{*}{LLaMA 7B} & 0 & 0.77\textsubscript{0.08} & 0.70\textsubscript{0.07} & 0.74\textsubscript{0.12} & 0.46\textsubscript{0.04} & 0.53\textsubscript{0.05} & 0.55\textsubscript{0.10} & 0.72\textsubscript{0.05} & 0.65\textsubscript{0.06} & 0.66\textsubscript{0.06} & 0.34\textsubscript{0.04} & 0.25\textsubscript{0.07} & 0.30\textsubscript{0.03} & \multirow{2}{*}{8/12}\\
& 2 & 0.72\textsubscript{0.17} & 0.83\textsubscript{0.07} & 0.83\textsubscript{0.11} & 0.38\textsubscript{0.21} & 0.76\textsubscript{0.06} & 0.73\textsubscript{0.13} & 0.61\textsubscript{0.24} & 0.75\textsubscript{0.08} & 0.72\textsubscript{0.13} & 0.29\textsubscript{0.10} & 0.38\textsubscript{0.07} & 0.38\textsubscript{0.11} \\
\multirow{2}{*}{LLaMA 13B} & 0 & 0.81\textsubscript{0.03} & 0.69\textsubscript{0.07} & 0.77\textsubscript{0.08} & 0.42\textsubscript{0.04} & 0.52\textsubscript{0.07} & 0.65\textsubscript{0.11} & 0.74\textsubscript{0.03} & 0.62\textsubscript{0.07} & 0.73\textsubscript{0.04} & 0.34\textsubscript{0.04} & 0.18\textsubscript{0.05} & 0.34\textsubscript{0.03} & \multirow{2}{*}{10/12}\\
& 2 & 0.75\textsubscript{0.17} & 0.83\textsubscript{0.08} & 0.82\textsubscript{0.14} & 0.38\textsubscript{0.17} & 0.75\textsubscript{0.06} & 0.80\textsubscript{0.12} & 0.68\textsubscript{0.15} & 0.75\textsubscript{0.07} & 0.80\textsubscript{0.08} & 0.35\textsubscript{0.09} & 0.36\textsubscript{0.08} & 0.42\textsubscript{0.09} \\
\multirow{2}{*}{LLaMA 30B} & 0 & 0.76\textsubscript{0.08} & 0.71\textsubscript{0.07} & 0.76\textsubscript{0.08} & 0.51\textsubscript{0.03} & 0.47\textsubscript{0.09} & 0.67\textsubscript{0.08} & 0.75\textsubscript{0.04} & 0.66\textsubscript{0.05} & 0.74\textsubscript{0.06} & 0.33\textsubscript{0.08} & 0.21\textsubscript{0.05} & 0.30\textsubscript{0.04} & \multirow{2}{*}{10/12}\\
& 2 & 0.78\textsubscript{0.17} & 0.81\textsubscript{0.10} & 0.83\textsubscript{0.14} & 0.43\textsubscript{0.19} & 0.76\textsubscript{0.09} & 0.80\textsubscript{0.09} & 0.65\textsubscript{0.22} & 0.74\textsubscript{0.13} & 0.78\textsubscript{0.07} & 0.34\textsubscript{0.11} & 0.41\textsubscript{0.08} & 0.43\textsubscript{0.13} \\
\multirow{2}{*}{LLaMA 65B} & 0 & 0.78\textsubscript{0.10} & 0.71\textsubscript{0.05} & 0.75\textsubscript{0.10} & 0.45\textsubscript{0.05} & 0.49\textsubscript{0.07} & 0.62\textsubscript{0.08} & 0.74\textsubscript{0.06} & 0.61\textsubscript{0.07} & 0.74\textsubscript{0.03} & 0.31\textsubscript{0.06} & 0.19\textsubscript{0.07} & 0.31\textsubscript{0.03} & \multirow{2}{*}{11/12}\\
& 2 & 0.82\textsubscript{0.17} & 0.84\textsubscript{0.09} & 0.87\textsubscript{0.13} & 0.45\textsubscript{0.17} & 0.78\textsubscript{0.06} & 0.80\textsubscript{0.13} & 0.68\textsubscript{0.20} & 0.78\textsubscript{0.08} & 0.82\textsubscript{0.05} & 0.38\textsubscript{0.08} & 0.38\textsubscript{0.09} & 0.45\textsubscript{0.11} \\
\midrule
\multirow{2}{*}{Falcon 1B} & 0 & 0.72\textsubscript{0.08} & 0.72\textsubscript{0.03} & 0.73\textsubscript{0.07} & 0.54\textsubscript{0.03} & 0.55\textsubscript{0.04} & 0.62\textsubscript{0.10} & 0.68\textsubscript{0.04} & 0.64\textsubscript{0.06} & 0.63\textsubscript{0.08} & 0.24\textsubscript{0.04} & 0.25\textsubscript{0.04} & 0.31\textsubscript{0.02} & \multirow{2}{*}{9/12}\\
& 2 & 0.65\textsubscript{0.17} & 0.77\textsubscript{0.10} & 0.71\textsubscript{0.17} & 0.36\textsubscript{0.25} & 0.72\textsubscript{0.05} & 0.74\textsubscript{0.14} & 0.52\textsubscript{0.17} & 0.72\textsubscript{0.08} & 0.77\textsubscript{0.08} & 0.26\textsubscript{0.09} & 0.33\textsubscript{0.06} & 0.33\textsubscript{0.06} \\
\multirow{2}{*}{Falcon 7B} & 0 & 0.72\textsubscript{0.09} & 0.68\textsubscript{0.05} & 0.73\textsubscript{0.08} & 0.50\textsubscript{0.06} & 0.51\textsubscript{0.13} & 0.66\textsubscript{0.06} & 0.75\textsubscript{0.06} & 0.64\textsubscript{0.03} & 0.72\textsubscript{0.06} & 0.31\textsubscript{0.04} & 0.21\textsubscript{0.07} & 0.29\textsubscript{0.03} & \multirow{2}{*}{10/12}\\
& 2 & 0.77\textsubscript{0.16} & 0.78\textsubscript{0.09} & 0.79\textsubscript{0.15} & 0.40\textsubscript{0.21} & 0.76\textsubscript{0.06} & 0.80\textsubscript{0.17} & 0.51\textsubscript{0.20} & 0.76\textsubscript{0.07} & 0.73\textsubscript{0.12} & 0.32\textsubscript{0.09} & 0.33\textsubscript{0.08} & 0.37\textsubscript{0.11} \\
\multirow{2}{*}{Falcon 40B} & 0 & 0.76\textsubscript{0.05} & 0.68\textsubscript{0.07} & 0.74\textsubscript{0.11} & 0.45\textsubscript{0.03} & 0.57\textsubscript{0.07} & 0.69\textsubscript{0.08} & 0.75\textsubscript{0.07} & 0.62\textsubscript{0.07} & 0.72\textsubscript{0.08} & 0.31\textsubscript{0.07} & 0.27\textsubscript{0.10} & 0.27\textsubscript{0.02} & \multirow{2}{*}{11/12}\\
& 2 & 0.79\textsubscript{0.17} & 0.81\textsubscript{0.09} & 0.87\textsubscript{0.13} & 0.42\textsubscript{0.15} & 0.83\textsubscript{0.06} & 0.85\textsubscript{0.12} & 0.64\textsubscript{0.23} & 0.79\textsubscript{0.09} & 0.80\textsubscript{0.06} & 0.36\textsubscript{0.07} & 0.41\textsubscript{0.06} & 0.45\textsubscript{0.06} \\
\midrule
\rowcolor{Gray}
& 0 & 0.70\textsubscript{0.12} & 0.59\textsubscript{0.08} & 0.62\textsubscript{0.16} & 0.35\textsubscript{0.04} & 0.29\textsubscript{0.09} & 0.21\textsubscript{0.22} & 0.68\textsubscript{0.04} & 0.45\textsubscript{0.10} & 0.41\textsubscript{0.23} & 0.30\textsubscript{0.06} & 0.13\textsubscript{0.05} & 0.15\textsubscript{0.16} & \\
\rowcolor{Gray}
\multirow{-2}{*}{Llama 2 7B} & 2 & 0.66\textsubscript{0.13} & 0.69\textsubscript{0.10} & 0.66\textsubscript{0.16} & 0.14\textsubscript{0.11} & 0.17\textsubscript{0.14} & 0.16\textsubscript{0.18} & 0.37\textsubscript{0.14} & 0.40\textsubscript{0.11} & 0.42\textsubscript{0.20} & 0.26\textsubscript{0.09} & 0.29\textsubscript{0.09} & 0.21\textsubscript{0.19} & \multirow{-2}{*}{5/12}\\
\multirow{2}{*}{Llama 2 13B} & 0 & 0.77\textsubscript{0.09} & 0.71\textsubscript{0.04} & 0.74\textsubscript{0.10} & 0.45\textsubscript{0.03} & 0.59\textsubscript{0.05} & 0.63\textsubscript{0.10} & 0.75\textsubscript{0.07} & 0.66\textsubscript{0.05} & 0.76\textsubscript{0.05} & 0.33\textsubscript{0.03} & 0.27\textsubscript{0.07} & 0.34\textsubscript{0.03} & \multirow{2}{*}{9/12}\\
& 2 & 0.79\textsubscript{0.17} & 0.82\textsubscript{0.10} & 0.88\textsubscript{0.09} & 0.40\textsubscript{0.15} & 0.79\textsubscript{0.06} & 0.83\textsubscript{0.11} & 0.70\textsubscript{0.15} & 0.76\textsubscript{0.09} & 0.81\textsubscript{0.05} & 0.32\textsubscript{0.09} & 0.35\textsubscript{0.11} & 0.46\textsubscript{0.11} \\
\multirow{2}{*}{Llama 2 70B} & 0 & 0.85\textsubscript{0.05} & 0.68\textsubscript{0.08} & 0.77\textsubscript{0.10} & 0.48\textsubscript{0.04} & 0.53\textsubscript{0.06} & 0.72\textsubscript{0.13} & 0.78\textsubscript{0.06} & 0.64\textsubscript{0.05} & 0.77\textsubscript{0.06} & 0.34\textsubscript{0.03} & 0.18\textsubscript{0.07} & 0.34\textsubscript{0.04} & \multirow{2}{*}{11/12}\\
& 2 & 0.83\textsubscript{0.14} & 0.83\textsubscript{0.08} & 0.88\textsubscript{0.13} & 0.46\textsubscript{0.15} & 0.79\textsubscript{0.10} & 0.84\textsubscript{0.12} & 0.76\textsubscript{0.14} & 0.78\textsubscript{0.09} & 0.81\textsubscript{0.09} & 0.41\textsubscript{0.07} & 0.41\textsubscript{0.10} & 0.51\textsubscript{0.06} \\
\midrule
\multicolumn{2}{l}{Highest mean, \%} & \textbf{37.5} & 32.5 & 30.0 & 5.0 & 12.5 & \textbf{82.5} & 32.5 & 10.0 & \textbf{57.5} & 25.0 & 15.0 & \textbf{60.0}\\
\multicolumn{2}{l}{Lowest std, \%} & 12.5 & \textbf{80.0} & 7.5 & 42.5 & \textbf{55.0} & 2.5 & 20.0 & \textbf{62.5} & 17.5 & 17.5 & 25.0 & \textbf{57.5} \\
\bottomrule
    \end{tabular}
    \caption{Evaluation of advanced prediction methods for all models  on 4 datasets in 0-shot and 2-shot with random demonstrations. Models that were removed from further evaluation are highlighted in gray.
    ``Calib.'' stands for the Calibration prompting method.}.
    \label{table:prediction_methods_full}
    \end{table*}
    
\section{Full Baseline Results}
\label{app:baseline_full}
\Cref{table:baseline_full} shows the results of evaluation of all \nmodels models in the default setting with a varying number of few-shot examples. 
These results illustrate that the template sensitivity issue is present in all models regardless of their size, and is not efficiently mitigated with the increase in the number of demonstrations.

\begin{table*}[]
    \centering
    \small
    \setlength{\tabcolsep}{3pt}
    \begin{tabular}{lcccccccccccc}
    \toprule
    \multirow{2}{*}{Model} & \multicolumn{3}{c}{SST-2} & \multicolumn{3}{c}{DBPedia} & \multicolumn{3}{c}{AGNews} & \multicolumn{3}{c}{TREC} \\
    \cmidrule(lr){2-4}\cmidrule(lr){5-7}\cmidrule(lr){8-10}\cmidrule(lr){11-13}
     & 0 & 2 & 4 & 0 & 2 & 4 & 0 & 2 & 4 & 0 & 2 & 4 \\
\midrule
GPT-J & 0.71\textsubscript{0.09} & 0.65\textsubscript{0.14} & 0.66\textsubscript{0.14} & 0.41\textsubscript{0.07} & 0.25\textsubscript{0.16} & 0.34\textsubscript{0.18} & 0.61\textsubscript{0.08} & 0.47\textsubscript{0.19} & 0.48\textsubscript{0.23} & 0.32\textsubscript{0.05} & 0.26\textsubscript{0.07} & 0.36\textsubscript{0.11} \\
GPT-NeoX & 0.71\textsubscript{0.08} & 0.69\textsubscript{0.15} & 0.82\textsubscript{0.12} & 0.48\textsubscript{0.04} & 0.32\textsubscript{0.19} & 0.37\textsubscript{0.21} & 0.67\textsubscript{0.06} & 0.52\textsubscript{0.22} & 0.48\textsubscript{0.22} & 0.30\textsubscript{0.08} & 0.31\textsubscript{0.08} & 0.40\textsubscript{0.10} \\
\midrule
OPT 6.7B & 0.79\textsubscript{0.07} & 0.67\textsubscript{0.16} & 0.80\textsubscript{0.14} & 0.46\textsubscript{0.04} & 0.27\textsubscript{0.14} & 0.33\textsubscript{0.18} & 0.59\textsubscript{0.08} & 0.45\textsubscript{0.17} & 0.47\textsubscript{0.22} & 0.24\textsubscript{0.04} & 0.27\textsubscript{0.08} & 0.34\textsubscript{0.10} \\
OPT 30B & 0.79\textsubscript{0.06} & 0.64\textsubscript{0.17} & 0.79\textsubscript{0.14} & 0.48\textsubscript{0.04} & 0.34\textsubscript{0.21} & 0.39\textsubscript{0.19} & 0.64\textsubscript{0.05} & 0.55\textsubscript{0.19} & 0.61\textsubscript{0.18} & 0.24\textsubscript{0.03} & 0.31\textsubscript{0.06} & 0.34\textsubscript{0.10} \\
OPT 66B & 0.73\textsubscript{0.12} & 0.65\textsubscript{0.15} & 0.84\textsubscript{0.13} & 0.41\textsubscript{0.03} & 0.34\textsubscript{0.16} & 0.40\textsubscript{0.16} & 0.64\textsubscript{0.07} & 0.45\textsubscript{0.17} & 0.53\textsubscript{0.19} & 0.26\textsubscript{0.03} & 0.28\textsubscript{0.07} & 0.33\textsubscript{0.09} \\
\midrule
BLOOM 1.7B & 0.68\textsubscript{0.11} & 0.66\textsubscript{0.12} & 0.67\textsubscript{0.13} & 0.47\textsubscript{0.03} & 0.27\textsubscript{0.19} & 0.31\textsubscript{0.20} & 0.61\textsubscript{0.08} & 0.43\textsubscript{0.19} & 0.42\textsubscript{0.19} & 0.27\textsubscript{0.04} & 0.31\textsubscript{0.09} & 0.36\textsubscript{0.11} \\
BLOOM 3B & 0.71\textsubscript{0.10} & 0.72\textsubscript{0.14} & 0.76\textsubscript{0.12} & 0.39\textsubscript{0.06} & 0.27\textsubscript{0.21} & 0.33\textsubscript{0.21} & 0.66\textsubscript{0.02} & 0.45\textsubscript{0.19} & 0.46\textsubscript{0.22} & 0.22\textsubscript{0.06} & 0.34\textsubscript{0.09} & 0.39\textsubscript{0.12} \\
BLOOM 7.1B & 0.72\textsubscript{0.09} & 0.69\textsubscript{0.15} & 0.74\textsubscript{0.15} & 0.44\textsubscript{0.05} & 0.26\textsubscript{0.18} & 0.32\textsubscript{0.21} & 0.64\textsubscript{0.06} & 0.43\textsubscript{0.17} & 0.41\textsubscript{0.21} & 0.35\textsubscript{0.07} & 0.33\textsubscript{0.08} & 0.38\textsubscript{0.10} \\
\midrule
Pythia 6.9B & 0.75\textsubscript{0.08} & 0.63\textsubscript{0.12} & 0.77\textsubscript{0.14} & 0.45\textsubscript{0.05} & 0.28\textsubscript{0.16} & 0.35\textsubscript{0.19} & 0.58\textsubscript{0.14} & 0.43\textsubscript{0.17} & 0.43\textsubscript{0.20} & 0.31\textsubscript{0.07} & 0.34\textsubscript{0.09} & 0.38\textsubscript{0.13} \\
Pythia 12B & 0.73\textsubscript{0.07} & 0.63\textsubscript{0.13} & 0.81\textsubscript{0.13} & 0.43\textsubscript{0.05} & 0.29\textsubscript{0.15} & 0.35\textsubscript{0.16} & 0.61\textsubscript{0.09} & 0.53\textsubscript{0.18} & 0.46\textsubscript{0.22} & 0.33\textsubscript{0.06} & 0.29\textsubscript{0.09} & 0.35\textsubscript{0.12} \\
\midrule
LLaMA 7B & 0.77\textsubscript{0.08} & 0.72\textsubscript{0.17} & 0.85\textsubscript{0.10} & 0.46\textsubscript{0.04} & 0.38\textsubscript{0.21} & 0.46\textsubscript{0.20} & 0.72\textsubscript{0.05} & 0.61\textsubscript{0.24} & 0.64\textsubscript{0.18} & 0.34\textsubscript{0.04} & 0.29\textsubscript{0.10} & 0.39\textsubscript{0.15} \\
LLaMA 13B & 0.81\textsubscript{0.03} & 0.75\textsubscript{0.17} & 0.86\textsubscript{0.14} & 0.42\textsubscript{0.04} & 0.38\textsubscript{0.17} & 0.52\textsubscript{0.11} & 0.74\textsubscript{0.03} & 0.68\textsubscript{0.15} & 0.74\textsubscript{0.13} & 0.34\textsubscript{0.04} & 0.35\textsubscript{0.09} & 0.42\textsubscript{0.14} \\
LLaMA 30B & 0.76\textsubscript{0.08} & 0.78\textsubscript{0.17} & 0.87\textsubscript{0.16} & 0.51\textsubscript{0.03} & 0.43\textsubscript{0.19} & 0.53\textsubscript{0.16} & 0.75\textsubscript{0.04} & 0.65\textsubscript{0.22} & 0.71\textsubscript{0.19} & 0.33\textsubscript{0.08} & 0.34\textsubscript{0.11} & 0.42\textsubscript{0.16} \\
LLaMA 65B & 0.78\textsubscript{0.10} & 0.82\textsubscript{0.17} & 0.92\textsubscript{0.10} & 0.45\textsubscript{0.05} & 0.45\textsubscript{0.17} & 0.52\textsubscript{0.14} & 0.74\textsubscript{0.06} & 0.68\textsubscript{0.20} & 0.71\textsubscript{0.17} & 0.31\textsubscript{0.06} & 0.38\textsubscript{0.08} & 0.47\textsubscript{0.09} \\
\midrule
Falcon 1B & 0.72\textsubscript{0.08} & 0.65\textsubscript{0.17} & 0.77\textsubscript{0.15} & 0.54\textsubscript{0.03} & 0.36\textsubscript{0.25} & 0.44\textsubscript{0.23} & 0.68\textsubscript{0.04} & 0.52\textsubscript{0.17} & 0.56\textsubscript{0.19} & 0.24\textsubscript{0.04} & 0.26\textsubscript{0.09} & 0.31\textsubscript{0.09} \\
Falcon 7B & 0.72\textsubscript{0.09} & 0.77\textsubscript{0.16} & 0.83\textsubscript{0.16} & 0.50\textsubscript{0.06} & 0.40\textsubscript{0.21} & 0.49\textsubscript{0.18} & 0.75\textsubscript{0.06} & 0.51\textsubscript{0.20} & 0.60\textsubscript{0.19} & 0.31\textsubscript{0.04} & 0.32\textsubscript{0.09} & 0.39\textsubscript{0.11} \\
Falcon 40B & 0.76\textsubscript{0.05} & 0.79\textsubscript{0.17} & 0.92\textsubscript{0.07} & 0.45\textsubscript{0.03} & 0.42\textsubscript{0.15} & 0.54\textsubscript{0.06} & 0.75\textsubscript{0.07} & 0.64\textsubscript{0.23} & 0.75\textsubscript{0.09} & 0.31\textsubscript{0.07} & 0.36\textsubscript{0.07} & 0.46\textsubscript{0.10} \\
\midrule
Llama 2 13B & 0.77\textsubscript{0.09} & 0.79\textsubscript{0.17} & 0.92\textsubscript{0.07} & 0.45\textsubscript{0.03} & 0.40\textsubscript{0.15} & 0.51\textsubscript{0.09} & 0.75\textsubscript{0.07} & 0.70\textsubscript{0.15} & 0.76\textsubscript{0.09} & 0.33\textsubscript{0.03} & 0.32\textsubscript{0.09} & 0.41\textsubscript{0.14} \\
Llama 2 70B & 0.85\textsubscript{0.05} & 0.83\textsubscript{0.14} & 0.92\textsubscript{0.09} & 0.48\textsubscript{0.04} & 0.46\textsubscript{0.15} & 0.60\textsubscript{0.05} & 0.78\textsubscript{0.06} & 0.76\textsubscript{0.14} & 0.82\textsubscript{0.05} & 0.34\textsubscript{0.03} & 0.41\textsubscript{0.07} & 0.51\textsubscript{0.06} \\
\bottomrule
\end{tabular}
\caption{Classification accuracy of all models on all datasets in the default setting. The results are aggregated over 10 templates for each set of random demonstrations, i.e., 10 runs for 0-shot and 30 runs for few-shot learning.}
\label{table:baseline_full}
\end{table*}

\section{Template Parts Analysis}
\label{app:template-parts}

\begin{figure*}[ht!]
\centering
\begin{subfigure}{\textwidth}
    \includegraphics[width=\linewidth]{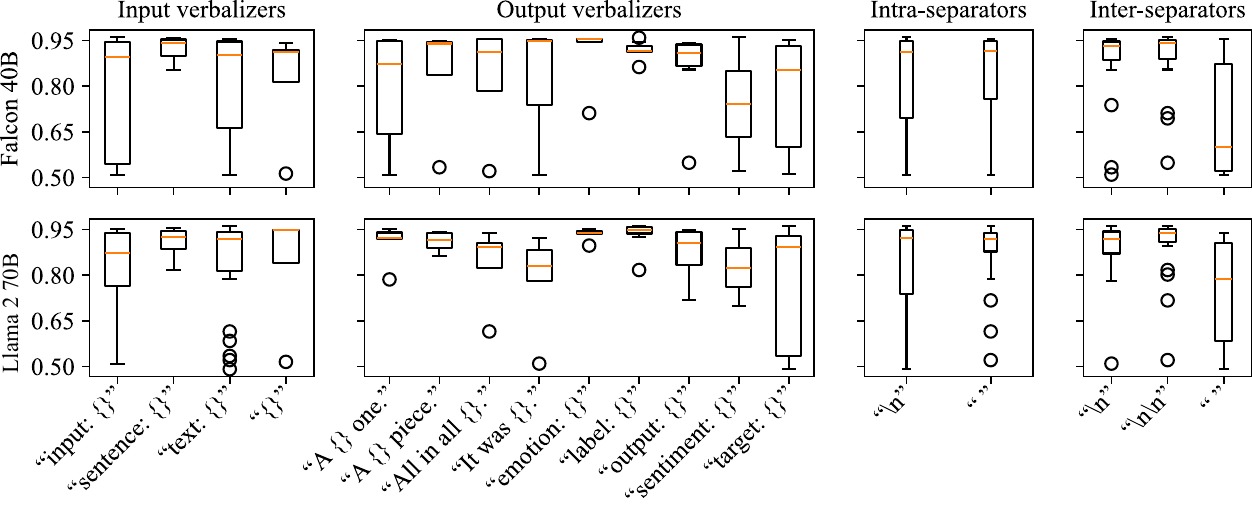}
    \label{fig:verbalisers_direct}
    \vspace{-12pt}
    \caption{\textsc{Direct} prediction method.}
\end{subfigure}

\vspace{10pt}
\begin{subfigure}{\textwidth}
    \includegraphics[width=\linewidth]{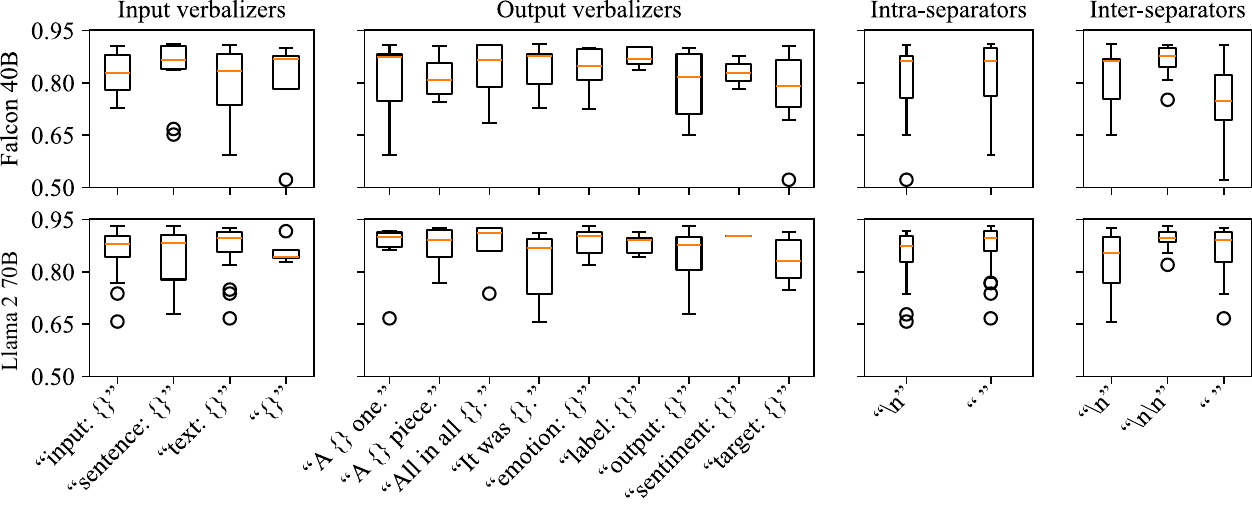}
    \label{fig:verbalisers_ch}
    \vspace{-12pt}
    \caption{\textsc{Channel} prediction method.}
\end{subfigure}

\vspace{10pt}
\begin{subfigure}{\textwidth}
    \includegraphics[width=\linewidth]{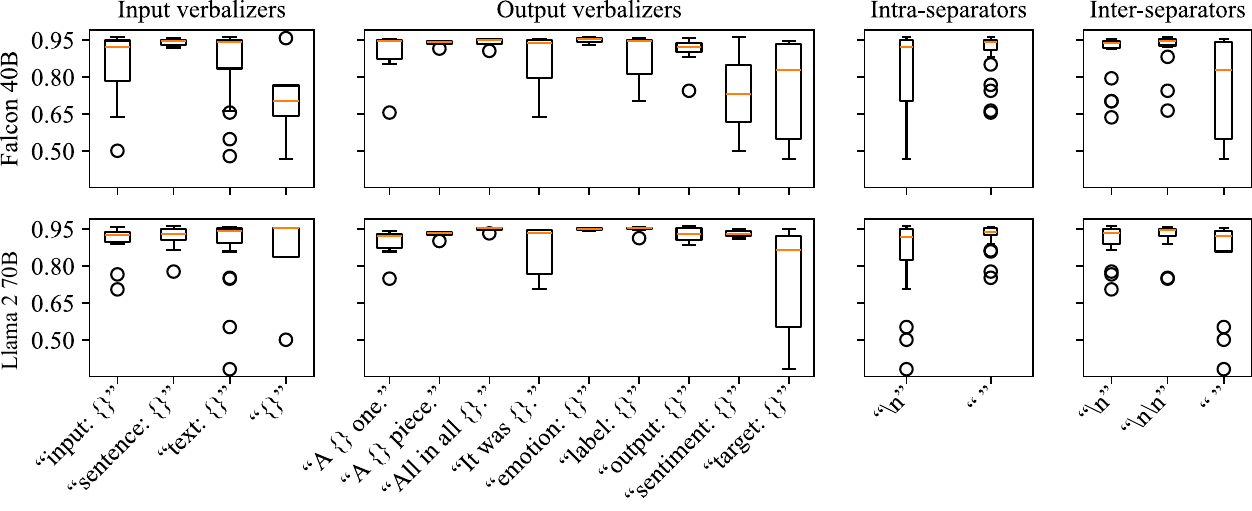}
    \label{fig:verbalisers_clb}
    \vspace{-12pt}
    \caption{\textsc{Calibration} prediction method.}
    \vspace{6pt}
\end{subfigure}
\caption{Accuracy for evaluation of templates with fixed parts on the \textsc{SST-2} dataset with \textsc{Random} 2-shot for all prediction methods.}
\label{fig:verbalisers}
\end{figure*}

In addition to studying prompt format sensitivity in general, we analyze how each part of a template impacts model performance.
For instance, it could be possible that the inclusion of a certain verbalizer in a template consistently leads to a decline in accuracy, irrespective of the other components.

To find that out, we decompose all templates into their parts and measure the distribution of scores for different variations of each component separately. 
The results presented in ~\Cref{fig:verbalisers} illustrate that even for state-of-the-art models, such as Llama 2 70B and Falcon 40B, many components exhibit high variance; also, the variance differs between two models.
In other words, even if a certain template yields good performance and low variance for a given setup, it is not guaranteed to work consistently well in other setups, and changing a single component could have detrimental effects.

Along with the non-transferability of whole templates, we notice that individual components also do not transfer both between models and prediction methods.
For instance, while ``It\ was\ \{\}'' ranks highest among output verbalizers for Llama 2 70B with the \textsc{Direct} prediction method, it is one of the worst for Falcon 40B.

Moreover, while a combination of best verbalizers is often a well-performing template, it is not necessarily the best one; the same is applicable for ``bad'' verbalizers too. 
For example, ``input: \{\}\textbackslash n sentiment: \{\}\textbackslash n\textbackslash n'' is the best template for Falcon 40B with the \textsc{Direct} method, even though ``sentiment: \{\}'' is one of the ``worst'' output verbalizers for that model. 

In summary, there is a complex interaction between the components of a template and their influence on model performance. 
We hypothesize that the transfer of both whole prompt templates and their parts is limited and requires further analysis.

\section{Prediction Methods}
\label{app:prediction_methods_full}
We provide the results of advanced prediction methods evaluation for all models in 0-shot and 2-shot setting with random demonstrations in~\Cref{table:prediction_methods_full}. 
We conclude from this comparison that neither of the advanced prediction strategies do not decrease prompt format sensitivity consistently across models and datasets.
Moreover, when accounting for the spread in accuracy scores caused by this issue, the advantages of these methods over \textsc{Direct} become less apparent.

\section{Reproduction of Results for Advanced Selection Methods}
\label{app:reproduction}
To evaluate the sensitivity of example selection methods to the template choice, we compare how the results reported in original works on these methods change when evaluated on a set of random templates instead of a predefined single one.
For an accurate reproduction of original setups, we evaluate both \textsc{z-ICL} and \textsc{ITM} using \textsc{Channel} prompting with corresponding templates from~\Cref{table:all_templates_sst2,table:all_templates_dbpedia}.

In~\Cref{table:selection_methods_native_itm,table:selection_methods_native_zicl}, we present the scores for a single template used in original implementations in the column ``Reproduced'', aggregating results over 5 example selection seeds. 
The ``Random'' column shows average scores for fixed demonstrations on a set of 10 random templates.

We observe that the results obtained in our code differ from the ones reported in the papers presenting both methods (the ``Paper'' column).
The cause of this discrepancy is presumably the difference in tokenization during preprocessing of the datasets.
Both methods use the same codebase with tokenization specific to GPT tokenizers, which results in a significant drop in quality for OPT and LLaMA models.
By contrast, our tokenization approach is more general and preserves the ICL performance of these models.

\begin{table*}[]
    \centering
    \begin{tabular}{lcccccc}
         \toprule
         \multirow{2}{*}{Model} & \multicolumn{3}{c}{SST 2} & \multicolumn{3}{c}{DBPedia 14} \\
    \cmidrule(lr){2-4}\cmidrule(lr){5-7}
     & Paper & Reproduced & Random & Paper & Reproduced & Random \\
\midrule
         GPT2-Large & 0.86\textsubscript{0.01} & 0.84\textsubscript{0.02} & 0.80\textsubscript{0.11} & 0.57\textsubscript{0.03} & 0.62\textsubscript{0.05} & 0.63\textsubscript{0.07} \\
         GPT2-XL & 0.83\textsubscript{0.04} & 0.73\textsubscript{0.13} & 0.75\textsubscript{0.12} & 0.59\textsubscript{0.03} & 0.61\textsubscript{0.02} & 0.62\textsubscript{0.09} \\
         GPT-J & 0.88\textsubscript{0.02} & 0.87\textsubscript{0.02} & 0.86\textsubscript{0.06} & 0.60\textsubscript{0.04} & 0.54\textsubscript{0.05} & 0.71\textsubscript{0.05} \\
         OPT-6.7B & 0.74\textsubscript{0.03} & 0.86\textsubscript{0.03} & 0.87\textsubscript{0.04} & 0.29\textsubscript{0.02} & 0.62\textsubscript{0.04} & 0.72\textsubscript{0.06} \\
         LLaMA-7B & 0.61\textsubscript{0.05} & 0.81\textsubscript{0.12} & 0.87\textsubscript{0.07} & 0.17\textsubscript{0.01} & 0.68\textsubscript{0.02} & 0.80\textsubscript{0.06} \\
         \bottomrule
    \end{tabular}
    \caption{Mean accuracy and standard deviation of 4-shot learning with \textsc{ITM} demonstrations using the \textsc{Channel} prediction method.}
    \label{table:selection_methods_native_itm}
\end{table*}

\begin{table}[]
    \centering
    \begin{tabular}{lcccc}
         \toprule
         Model & Paper & Reproduced & Random \\
\midrule
         GPT-J &  0.83\textsubscript{0.02} & 0.86\textsubscript{0.00} & 
         0.61\textsubscript{0.06} \\
         GPT-NeoX &  0.79\textsubscript{0.00} & 0.83\textsubscript{0.00} & 0.80\textsubscript{0.04} \\
         \bottomrule
    \end{tabular}
    \caption{Mean accuracy and standard deviation of 4-shot learning with  \textsc{z-ICL} demonstrations on SST-2 dataset using the \textsc{Channel} prediction method.}
    \label{table:selection_methods_native_zicl}
\end{table}

From these results, we conclude that both methods are not robust to the template choice, as the mean performance decreases for multiple models while the standard deviation across seeds increases.
Therefore, the gains from advanced example selection methods are caused to a certain degree by the choice of a proper prompt format rather than the retrieved demonstrations.

\section{Example Selection Methods}
\label{app:selection_methods_full}
Full results of evaluation of demonstration selection techniques in 4-shot learning using the \textsc{Direct} prediction method are presented in~\Cref{table:selection_methods_full}.
\begin{table*}[h!]
    \centering
    \small
    \setlength{\tabcolsep}{2.5pt}
    \begin{tabular}{lccccccccccccc}
    \toprule
    \multirow{2}{*}{Model} & 
    \multicolumn{3}{c}{SST-2} & \multicolumn{3}{c}{DBPedia} & \multicolumn{3}{c}{AGNews} & \multicolumn{3}{c}{TREC} \\
    \cmidrule(lr){2-4}\cmidrule(lr){5-7}\cmidrule(lr){8-10}\cmidrule(lr){11-13}&
    Random & ITM & z-ICL & Random & ITM & z-ICL & Random & ITM & z-ICL & Random & ITM & z-ICL \\
\midrule
GPT-J & 0.66\textsubscript{0.14} & 0.73\textsubscript{0.16} & 0.66\textsubscript{0.05} & 0.34\textsubscript{0.18} & 0.46\textsubscript{0.17} & 0.24\textsubscript{0.10} & 0.48\textsubscript{0.23} & 0.59\textsubscript{0.12} & 0.45\textsubscript{0.11} & 0.36\textsubscript{0.11} & 0.25\textsubscript{0.10} & 0.21\textsubscript{0.04} \\
GPT-NeoX & 0.82\textsubscript{0.12} & 0.73\textsubscript{0.17} & 0.77\textsubscript{0.09} & 0.37\textsubscript{0.21} & 0.45\textsubscript{0.19} & 0.25\textsubscript{0.11} & 0.48\textsubscript{0.22} & 0.58\textsubscript{0.17} & 0.46\textsubscript{0.12} & 0.40\textsubscript{0.10} & 0.28\textsubscript{0.09} & 0.25\textsubscript{0.07} \\
\midrule
OPT 6.7B & 0.80\textsubscript{0.14} & 0.77\textsubscript{0.18} & 0.72\textsubscript{0.09} & 0.33\textsubscript{0.18} & 0.46\textsubscript{0.17} & 0.24\textsubscript{0.11} & 0.47\textsubscript{0.22} & 0.49\textsubscript{0.17} & 0.45\textsubscript{0.14} & 0.34\textsubscript{0.10} & 0.23\textsubscript{0.07} & 0.22\textsubscript{0.05} \\
OPT 30B & 0.79\textsubscript{0.14} & 0.80\textsubscript{0.16} & 0.79\textsubscript{0.12} & 0.39\textsubscript{0.19} & 0.50\textsubscript{0.16} & 0.27\textsubscript{0.11} & 0.61\textsubscript{0.18} & 0.67\textsubscript{0.14} & 0.54\textsubscript{0.13} & 0.34\textsubscript{0.10} & 0.25\textsubscript{0.08} & 0.25\textsubscript{0.06} \\
OPT 66B & 0.84\textsubscript{0.13} & 0.83\textsubscript{0.15} & 0.75\textsubscript{0.08} & 0.40\textsubscript{0.16} & 0.48\textsubscript{0.14} & 0.25\textsubscript{0.10} & 0.53\textsubscript{0.19} & 0.61\textsubscript{0.15} & 0.48\textsubscript{0.12} & 0.33\textsubscript{0.09} & 0.25\textsubscript{0.09} & 0.21\textsubscript{0.04} \\
\midrule
BLOOM 1.7B & 0.67\textsubscript{0.13} & 0.69\textsubscript{0.14} & 0.64\textsubscript{0.09} & 0.31\textsubscript{0.20} & 0.39\textsubscript{0.19} & 0.20\textsubscript{0.07} & 0.42\textsubscript{0.19} & 0.46\textsubscript{0.16} & 0.45\textsubscript{0.08} & 0.36\textsubscript{0.11} & 0.26\textsubscript{0.07} & 0.19\textsubscript{0.06} \\
BLOOM 3B & 0.76\textsubscript{0.12} & 0.72\textsubscript{0.15} & 0.62\textsubscript{0.07} & 0.33\textsubscript{0.21} & 0.43\textsubscript{0.18} & 0.19\textsubscript{0.08} & 0.46\textsubscript{0.22} & 0.50\textsubscript{0.16} & 0.43\textsubscript{0.12} & 0.39\textsubscript{0.12} & 0.29\textsubscript{0.10} & 0.23\textsubscript{0.03} \\
BLOOM 7.1B & 0.74\textsubscript{0.15} & 0.71\textsubscript{0.15} & 0.62\textsubscript{0.08} & 0.32\textsubscript{0.21} & 0.44\textsubscript{0.19} & 0.20\textsubscript{0.08} & 0.41\textsubscript{0.21} & 0.50\textsubscript{0.17} & 0.42\textsubscript{0.10} & 0.38\textsubscript{0.10} & 0.32\textsubscript{0.08} & 0.21\textsubscript{0.05} \\
\midrule
Pythia 6.9B & 0.77\textsubscript{0.14} & 0.72\textsubscript{0.17} & 0.70\textsubscript{0.10} & 0.35\textsubscript{0.19} & 0.47\textsubscript{0.17} & 0.22\textsubscript{0.10} & 0.43\textsubscript{0.20} & 0.56\textsubscript{0.16} & 0.46\textsubscript{0.12} & 0.38\textsubscript{0.13} & 0.31\textsubscript{0.08} & 0.22\textsubscript{0.05} \\
Pythia 12B & 0.81\textsubscript{0.13} & 0.72\textsubscript{0.18} & 0.79\textsubscript{0.10} & 0.35\textsubscript{0.16} & 0.46\textsubscript{0.14} & 0.24\textsubscript{0.10} & 0.46\textsubscript{0.22} & 0.57\textsubscript{0.16} & 0.44\textsubscript{0.13} & 0.35\textsubscript{0.12} & 0.30\textsubscript{0.09} & 0.22\textsubscript{0.04} \\
\midrule
LLaMA 7B & 0.85\textsubscript{0.10} & 0.84\textsubscript{0.16} & 0.70\textsubscript{0.09} & 0.46\textsubscript{0.20} & 0.51\textsubscript{0.10} & 0.27\textsubscript{0.08} & 0.64\textsubscript{0.18} & 0.66\textsubscript{0.15} & 0.52\textsubscript{0.13} & 0.39\textsubscript{0.15} & 0.28\textsubscript{0.07} & 0.27\textsubscript{0.05} \\
LLaMA 13B & 0.86\textsubscript{0.14} & 0.85\textsubscript{0.13} & 0.73\textsubscript{0.11} & 0.52\textsubscript{0.11} & 0.52\textsubscript{0.10} & 0.31\textsubscript{0.06} & 0.74\textsubscript{0.13} & 0.77\textsubscript{0.08} & 0.55\textsubscript{0.08} & 0.42\textsubscript{0.14} & 0.32\textsubscript{0.13} & 0.29\textsubscript{0.04} \\
LLaMA 30B & 0.87\textsubscript{0.16} & 0.88\textsubscript{0.11} & 0.67\textsubscript{0.08} & 0.53\textsubscript{0.16} & 0.57\textsubscript{0.15} & 0.27\textsubscript{0.06} & 0.71\textsubscript{0.19} & 0.77\textsubscript{0.08} & 0.45\textsubscript{0.07} & 0.42\textsubscript{0.16} & 0.30\textsubscript{0.11} & 0.29\textsubscript{0.07} \\
LLaMA 65B & 0.92\textsubscript{0.10} & 0.91\textsubscript{0.08} & 0.73\textsubscript{0.10} & 0.52\textsubscript{0.14} & 0.51\textsubscript{0.13} & 0.30\textsubscript{0.06} & 0.71\textsubscript{0.17} & 0.84\textsubscript{0.07} & 0.57\textsubscript{0.08} & 0.47\textsubscript{0.09} & 0.36\textsubscript{0.09} & 0.34\textsubscript{0.05} \\
\midrule
Falcon 1B & 0.77\textsubscript{0.15} & 0.77\textsubscript{0.15} & 0.71\textsubscript{0.09} & 0.44\textsubscript{0.23} & 0.59\textsubscript{0.18} & 0.23\textsubscript{0.06} & 0.56\textsubscript{0.19} & 0.58\textsubscript{0.15} & 0.46\textsubscript{0.10} & 0.31\textsubscript{0.09} & 0.25\textsubscript{0.10} & 0.19\textsubscript{0.04} \\
Falcon 7B & 0.83\textsubscript{0.16} & 0.82\textsubscript{0.16} & 0.68\textsubscript{0.11} & 0.49\textsubscript{0.18} & 0.58\textsubscript{0.09} & 0.27\textsubscript{0.09} & 0.60\textsubscript{0.19} & 0.63\textsubscript{0.15} & 0.52\textsubscript{0.12} & 0.39\textsubscript{0.11} & 0.29\textsubscript{0.09} & 0.25\textsubscript{0.08} \\
Falcon 40B & 0.92\textsubscript{0.07} & 0.92\textsubscript{0.09} & 0.75\textsubscript{0.11} & 0.54\textsubscript{0.06} & 0.54\textsubscript{0.07} & 0.28\textsubscript{0.08} & 0.75\textsubscript{0.09} & 0.80\textsubscript{0.06} & 0.55\textsubscript{0.11} & 0.46\textsubscript{0.10} & 0.37\textsubscript{0.11} & 0.26\textsubscript{0.08} \\
\midrule
Llama 2 13B & 0.92\textsubscript{0.07} & 0.86\textsubscript{0.14} & 0.75\textsubscript{0.07} & 0.51\textsubscript{0.09} & 0.53\textsubscript{0.09} & 0.25\textsubscript{0.06} & 0.76\textsubscript{0.09} & 0.82\textsubscript{0.05} & 0.54\textsubscript{0.08} & 0.41\textsubscript{0.14} & 0.34\textsubscript{0.11} & 0.29\textsubscript{0.04} \\
Llama 2 70B & 0.92\textsubscript{0.09} & 0.91\textsubscript{0.07} & 0.75\textsubscript{0.09} & 0.60\textsubscript{0.05} & 0.59\textsubscript{0.09} & 0.28\textsubscript{0.05} & 0.82\textsubscript{0.05} & 0.85\textsubscript{0.05} & 0.60\textsubscript{0.07} & 0.51\textsubscript{0.06} & 0.39\textsubscript{0.10} & 0.34\textsubscript{0.03} \\
\midrule
Highest mean, \% & 75.0 & 25.0 & 0.0 & 20.0 & 80.0 & 0.0 & 0.0 & 100.0 & 0.0 & 100.0 & 0.0 & 0.0\\
Lowest std, \% & 10.0 & 10.0 & 80.0 & 5.0 & 5.0 & 90.0 & 5.0 & 20.0 & 75.0 & 0.0 & 0.0 & 100.0\\
\bottomrule
    \end{tabular}
    \caption{Evaluation of advanced selection methods for all \nmodels models using \textsc{Direct} prediction method in 4-shot. Results are aggregated over 10 random templates for each of the three demonstrations selection seeds.}.
    \label{table:selection_methods_full}
    \vspace{12pt}
    \end{table*}
The results highlight that advanced example selection techniques often perform comparably to the random choice baseline when evaluated on multiple templates. 
One might argue that the prompt format choice is inseparable from the method itself and thus such a comparison is invalid. 
However, since the best-performing formats do not transfer between models or demonstration sets of different sizes selected with the same method, a proper evaluation would require finding the best template for each setup.
This procedure both is computationally expensive and difficult to accomplish, as authors of example selection methods frequently omit the description of their format selection algorithm.

\section{Accuracy as a Function of Template Rank}
\label{sec:templates-percent}
\begin{figure*}[b!]
    \centering
    \includegraphics[width=\textwidth]{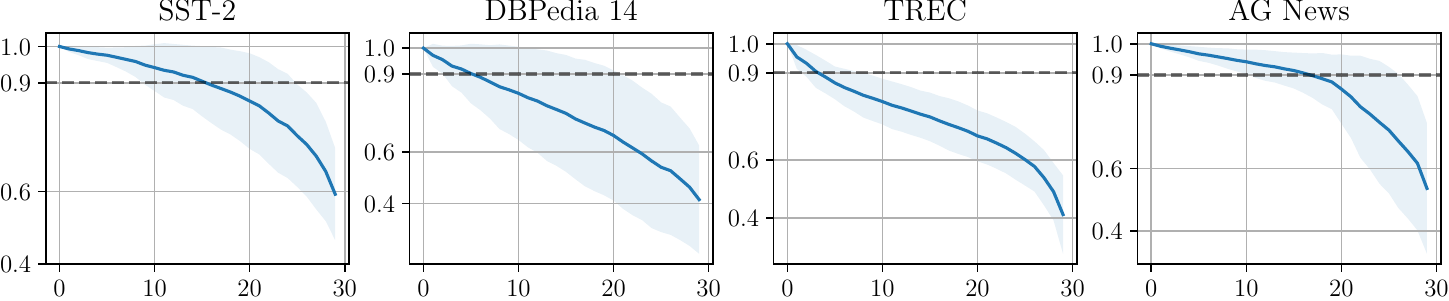}
    \caption{Relative quality of templates sorted by their classification accuracy. The shaded area indicates the standard deviation across \nmodels models.
    }
    \label{fig:accuracy-deterioration}
\end{figure*}

We plot the dependence of accuracy on the rank of the template in~\Cref{fig:accuracy-deterioration}. 
The results are aggregated across \nmodels models.
Each model was evaluated on 30 random templates with the \textsc{Direct} prediction method and the same set of 2 randomly selected demonstrations.
We observe that for SST-2 and AGNews datasets, the mean quality of the tenth-best template is within 0.9 of the best template score, which we consider a successful transfer.
Despite the more rapid decay for DBPedia and TREC, taking variation across models into account, we still count first 10 formats as performing on par with the best one.

\section{Transfer Evaluation with Spearman Rank Correlation}
\label{app:transfer-spearman}
\begin{figure*}[h!]
    \centering
    \includegraphics[width=\textwidth]{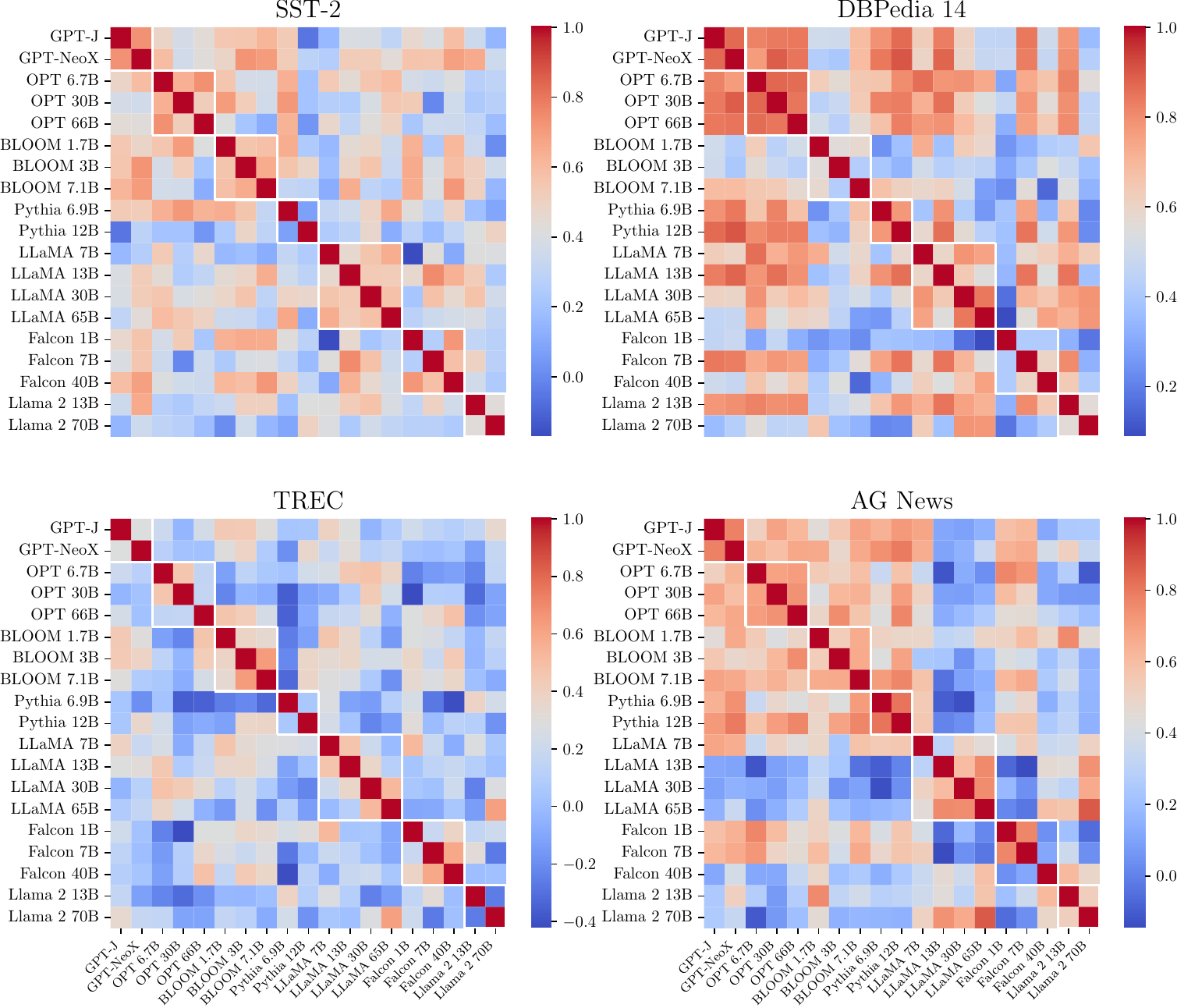}
    \caption{Spearman rank correlation over 30 templates for models evaluated in the \textsc{Direct}-\textsc{Random}-2-shot setting.}
    \label{fig:transfer-spearman}
\end{figure*}

One of the possible means to evaluate template transfer is to calculate the Spearman rank correlation between scores of all templates.
As can be seen from~\Cref{fig:transfer-spearman}, this method yields higher correlations than IoU over 10 best formats, but the capacity for transfer is still far from perfect (for example, for SST-2 and TREC datasets).

\section{IoU Transfer For All Datasets}
\label{sec:transfer-models}

\begin{figure*}[h!]
    \centering
    \includegraphics[width=\textwidth]{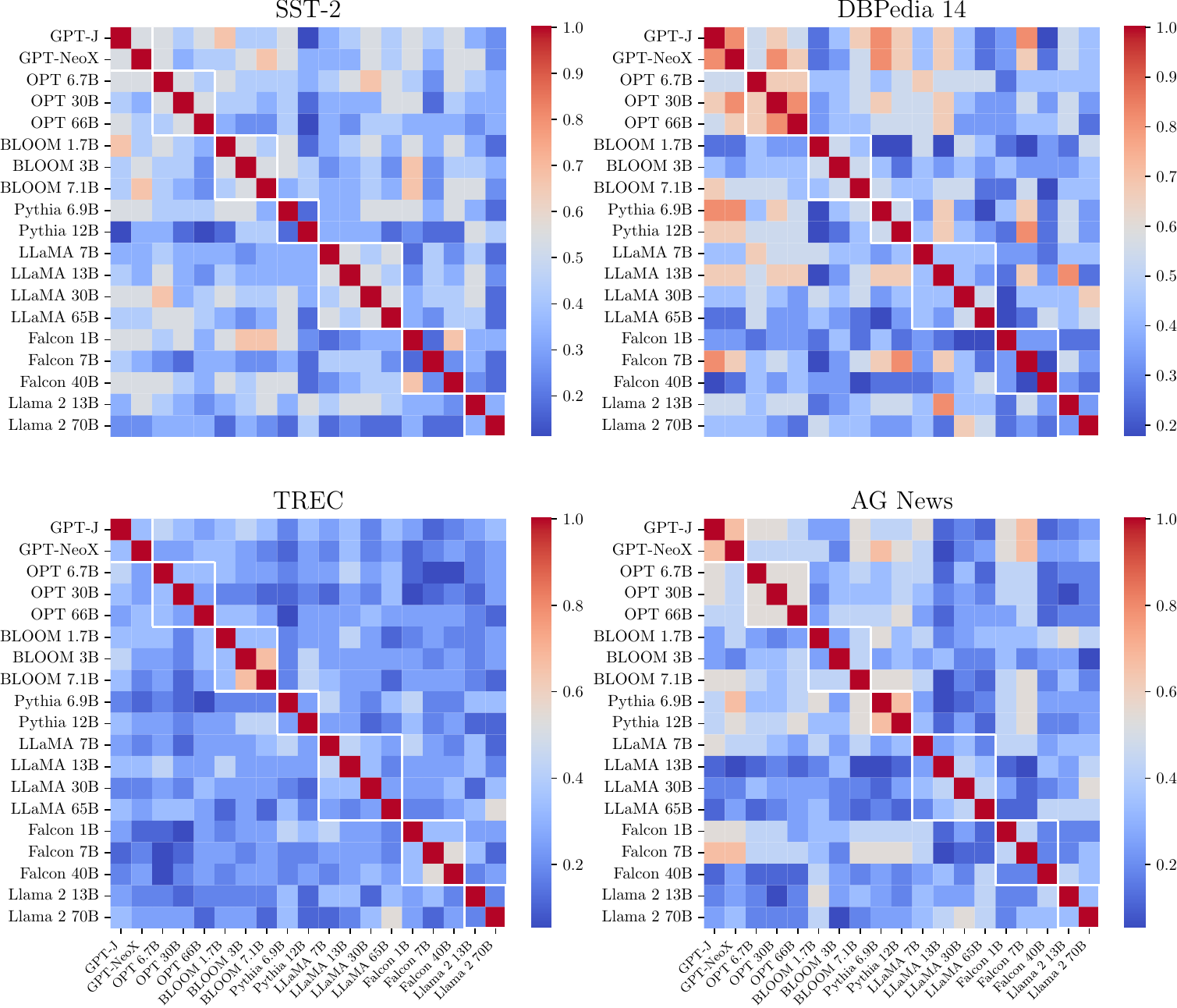}
    \caption{IoU of top-10 templates for all models evaluated in \textsc{Direct}-\textsc{Random}-2-shot setting.}
    \label{fig:transfer-iou-full}
    \vspace{-20pt}
\end{figure*}

Similarly to~\Cref{fig:transfer-models-iou-main}, we provide Intersection-over-Union of 10 best prompt formats for all \nmodels models and all datasets explored in our work in~\Cref{fig:transfer-iou-full}.
These heatmaps illustrate that the transfer of best-performing templates between models is remarkably low for all datasets.

\section{Additional Results For Template Ensembles}
\label{app:ensembles-full}

\Cref{table:full-ensembles-sst2,table:full-ensembles-trec} show the results of Template Ensembles evaluation on a broader set of models and datasets. 
For most setups, ensemble of size 5 exhibit better performance than a single template.

\begin{table*}[h!]
\centering
\begin{tabular}{lcccccc}
\toprule
\multirow{2}{*}{Model} & \multicolumn{2}{c}{Direct} & \multicolumn{2}{c}{Channel} & \multicolumn{2}{c}{Calibration} \\
           \cmidrule(lr){2-3}\cmidrule(lr){4-5}\cmidrule(lr){6-7}
     &  Single &      Ensemble &  Single &  Ensemble & Single &   Ensemble \\
\midrule
LLaMA 7B   &  0.73\textsubscript{0.17} &  0.81\textsubscript{0.16} &  0.83\textsubscript{0.06} &  0.89\textsubscript{0.03} &   0.82\textsubscript{0.13} &  \textbf{0.92}\textsubscript{0.02} \\
LLaMA 13B  &  0.76\textsubscript{0.17} &  0.82\textsubscript{0.15} &  0.83\textsubscript{0.08} &  \textbf{0.89}\textsubscript{0.03} &   0.78\textsubscript{0.18} &  0.87\textsubscript{0.08} \\
LLaMA 30B &  0.79\textsubscript{0.15} &  0.86\textsubscript{0.08} &  0.84\textsubscript{0.08} &  0.87\textsubscript{0.04} &   0.82\textsubscript{0.16} &  \textbf{0.93}\textsubscript{0.01} \\
LLaMA 65B  &  0.86\textsubscript{0.13} &  \textbf{0.95}\textsubscript{0.00} &  0.85\textsubscript{0.07} &  0.90\textsubscript{0.02} &   0.89\textsubscript{0.12} &  \textbf{0.95}\textsubscript{0.01} \\
\midrule
Llama 2 13B &   0.79\textsubscript{0.17}  & 0.85\textsubscript{0.09} &   0.82\textsubscript{0.10}  & 0.90\textsubscript{0.02} &    0.88\textsubscript{0.09}  & \textbf{0.93}\textsubscript{0.03} \\
Llama 2 70B &   0.83\textsubscript{0.14}  & \textbf{0.95}\textsubscript{0.01} &   0.83\textsubscript{0.08}  & 0.92\textsubscript{0.01} &    0.88\textsubscript{0.13}  & 0.94\textsubscript{0.03} \\
\midrule
Falcon 1B &   0.65\textsubscript{0.17}  & 0.74\textsubscript{0.07} &   0.77\textsubscript{0.10}  & \textbf{0.89}\textsubscript{0.01} &    0.71\textsubscript{0.17}  & \textbf{0.89}\textsubscript{0.01} \\
Falcon 7B &   0.77\textsubscript{0.16}  & 0.81\textsubscript{0.00} &   0.78\textsubscript{0.09}  & 0.90\textsubscript{0.00} &    0.79\textsubscript{0.15}  & \textbf{0.93}\textsubscript{0.02} \\
Falcon 40B &   0.79\textsubscript{0.17}  & 0.93\textsubscript{0.01} &   0.81\textsubscript{0.09}  & 0.91\textsubscript{0.01} &    0.87\textsubscript{0.13}  & \textbf{0.95}\textsubscript{0.00} \\
\bottomrule
\end{tabular}
\caption{Comparison of 2-shot learning performance on the SST-2 dataset using ensembles of 5 templates and a single template. Results are averaged over 5 random seeds.}
\label{table:full-ensembles-sst2}
\vspace{-10pt}
\end{table*}

\begin{table*}[h!]
\centering
\begin{tabular}{lcccccc}
\toprule
\multirow{2}{*}{Model} & \multicolumn{2}{c}{Direct} & \multicolumn{2}{c}{Channel} & \multicolumn{2}{c}{Calibration} \\
           \cmidrule(lr){2-3}\cmidrule(lr){4-5}\cmidrule(lr){6-7}
     &  Single &      Ensemble &  Single &  Ensemble & Single &   Ensemble \\
\midrule
LLaMA 7B &  0.29\textsubscript{0.10}  & 0.39\textsubscript{0.04} &  0.38\textsubscript{0.07}  & 0.48\textsubscript{0.10} &  0.38\textsubscript{0.11}  & \textbf{0.53}\textsubscript{0.05} \\
LLaMA 13B &  0.35\textsubscript{0.09}  & 0.43\textsubscript{0.09} &  0.36\textsubscript{0.08}  & 0.46\textsubscript{0.03} &  0.42\textsubscript{0.09}  & \textbf{0.55}\textsubscript{0.01} \\
LLaMA 30B &  0.34\textsubscript{0.11}  & 0.38\textsubscript{0.04} &  0.41\textsubscript{0.08}  & 0.55\textsubscript{0.02} &  0.43\textsubscript{0.13}  & \textbf{0.53}\textsubscript{0.03} \\
LLaMA 65B & 0.38\textsubscript{0.08} & 0.48\textsubscript{0.07}& 0.38\textsubscript{0.09} & \textbf{0.56}\textsubscript{0.01} & 0.45\textsubscript{0.11} & 0.55 \textsubscript{0.04}\\
\midrule
Llama 2 13B &  0.32\textsubscript{0.09}  & 0.45\textsubscript{0.11} &  0.35\textsubscript{0.11}  & 0.45\textsubscript{0.02} &  0.46\textsubscript{0.11}  & \textbf{0.58}\textsubscript{0.05} \\
Llama 2 70B &  0.41\textsubscript{0.07}  & 0.42\textsubscript{0.09} &  0.41\textsubscript{0.10}  & 0.46\textsubscript{0.03} &  0.51\textsubscript{0.06}  & \textbf{0.57}\textsubscript{0.02} \\
\midrule
Falcon 1B &  0.26\textsubscript{0.09}  & 0.33\textsubscript{0.06} &  0.33\textsubscript{0.06}  & 0.43\textsubscript{0.06} &  0.33\textsubscript{0.06}  & \textbf{0.44}\textsubscript{0.01} \\
Falcon 7B &  0.32\textsubscript{0.09}  & 0.36\textsubscript{0.05} &  0.33\textsubscript{0.08}  & \textbf{0.49}\textsubscript{0.12} &  0.37\textsubscript{0.11}  & 0.48\textsubscript{0.06} \\
Falcon 40B &  0.36\textsubscript{0.07}  & 0.39\textsubscript{0.06} &  0.41\textsubscript{0.06}  & 0.52\textsubscript{0.03} &  0.45\textsubscript{0.06}  & \textbf{0.53}\textsubscript{0.05} \\

\bottomrule
\end{tabular}

\caption{Comparison of 2-shot learning performance on the TREC dataset using ensembles of 5 templates and a single template. Results are averaged over 5 random seeds.}
\label{table:full-ensembles-trec}
\vspace{-10pt}
\end{table*}

\section{Evaluation of Instruct Models}
\label{app:instruct_res}
We validate that our findings hold true even for the latest instruction-tuned models, such as Llama 3 8B Instruct and Mistral v0.3 7B Instruct. 
First, as we show in~\Cref{table:baseline_instructs}, in the baseline setting, an increase in number of demonstrations generally leads to a better performance of instruct models but does not significantly decrease variance of their final scores, similarly to what we observe in base models.

Second,~\Cref{table:prediction_methods_instruct} demonstrates that, after adjusting to template robustness, the default \textsc{Direct} prediction method performs on par with more advanced methods, e.g. \textsc{Channel} and \textsc{Calibration}, while sometimes having a less variance of the model's scores.

Finally, we observe that the examples retrieved by \textsc{z-ICL} method turn out to be consistently worse than two other methods. 
This is also in line with our observation for base models.
However, in contrast to our main results, where there was no evident winner between two other methods, the demonstrations selected with \textsc{ITM} method turn out to be slightly more robust to template choice for instruction-tuned models, as we show in~\Cref{table:selection_methods_instruct}.

\begin{table*}[]
    \centering
    \small
    \setlength{\tabcolsep}{1.75pt}
    \begin{tabular}{lcccccccccccc}
    \toprule
    \multirow{2}{*}{Model} & \multicolumn{3}{c}{SST-2} & \multicolumn{3}{c}{DBPedia} & \multicolumn{3}{c}{AGNews} & \multicolumn{3}{c}{TREC} \\
    \cmidrule(lr){2-4}\cmidrule(lr){5-7}\cmidrule(lr){8-10}\cmidrule(lr){11-13}
     & 0 & 2 & 4 & 0 & 2 & 4 & 0 & 2 & 4 & 0 & 2 & 4 \\
     \midrule
Mistral v0.3 7B Instruct & 0.83\textsubscript{0.07} & 0.90\textsubscript{0.09} & 0.91\textsubscript{0.12} & 0.50\textsubscript{0.06} & 0.64\textsubscript{0.05} & 0.65\textsubscript{0.02} & 0.34\textsubscript{0.05} & 0.40\textsubscript{0.08} & 0.52\textsubscript{0.08} & 0.71\textsubscript{0.07} & 0.82\textsubscript{0.07} & 0.76\textsubscript{0.08} \\
Llama 3 8B Instruct & 0.77\textsubscript{0.12} & 0.87\textsubscript{0.13} & 0.91\textsubscript{0.10} & 0.48\textsubscript{0.02} & 0.55\textsubscript{0.13} & 0.63\textsubscript{0.05} & 0.24\textsubscript{0.06} & 0.39\textsubscript{0.09} & 0.48\textsubscript{0.11} & 0.77\textsubscript{0.06} & 0.77\textsubscript{0.18} & 0.78\textsubscript{0.10} \\
\bottomrule
\end{tabular}

\caption{Classification accuracy of instruction-tuned models on all datasets in the default setting. The results are averaged over
10 templates for each set of random demonstrations (3 sets in total), i.e., 10 runs for 0-shot and 30 runs for few-shot learning.}
\label{table:baseline_instructs}
\vspace{-10pt}
\end{table*}

\begin{table*}
\centering
    \small
    \setlength{\tabcolsep}{1.55pt}
    \begin{tabular}{lccccccccccccc}
    \toprule
    \multirow{2}{*}{Model} & \multirow{2}{*}{N} & \multicolumn{3}{c}{SST-2} & \multicolumn{3}{c}{DBPedia} & \multicolumn{3}{c}{AGNews} & \multicolumn{3}{c}{TREC} \\
    \cmidrule(lr){3-5}\cmidrule(lr){6-8}\cmidrule(lr){9-11}\cmidrule(lr){12-14}
     & & Direct & Channel & Calib. & Direct & Channel & Calib. & Direct & Channel & Calib. & Direct & Channel & Calib. \\
     \midrule
\multirow{2}{*}{Mistral v0.3 7B Instruct} & 0 & 0.83\textsubscript{0.07} & 0.73\textsubscript{0.05} & 0.83\textsubscript{0.10} & 0.50\textsubscript{0.06} & 0.47\textsubscript{0.08} & 0.75\textsubscript{0.09} & 0.34\textsubscript{0.05} & 0.19\textsubscript{0.08} & 0.30\textsubscript{0.07} & 0.71\textsubscript{0.07} & 0.59\textsubscript{0.06} & 0.68\textsubscript{0.06} \\
& 2 & 0.90\textsubscript{0.09} & 0.83\textsubscript{0.09} & 0.90\textsubscript{0.12} & 0.64\textsubscript{0.05} & 0.75\textsubscript{0.04} & 0.91\textsubscript{0.07} & 0.40\textsubscript{0.08} & 0.41\textsubscript{0.06} & 0.43\textsubscript{0.06} & 0.82\textsubscript{0.07} & 0.80\textsubscript{0.02} & 0.80\textsubscript{0.04} \\
\multirow{2}{*}{Llama 3 8B Instruct} & 0 & 0.77\textsubscript{0.12} & 0.66\textsubscript{0.04} & 0.74\textsubscript{0.14} & 0.48\textsubscript{0.02} & 0.48\textsubscript{0.05} & 0.72\textsubscript{0.08} & 0.24\textsubscript{0.06} & 0.18\textsubscript{0.08} & 0.22\textsubscript{0.04} & 0.77\textsubscript{0.06} & 0.57\textsubscript{0.06} & 0.76\textsubscript{0.06} \\
& 2 & 0.87\textsubscript{0.13} & 0.78\textsubscript{0.10} & 0.90\textsubscript{0.09} & 0.55\textsubscript{0.13} & 0.72\textsubscript{0.04} & 0.85\textsubscript{0.11} & 0.39\textsubscript{0.09} & 0.36\textsubscript{0.09} & 0.44\textsubscript{0.06} & 0.77\textsubscript{0.18} & 0.72\textsubscript{0.10} & 0.80\textsubscript{0.08} \\
\bottomrule
    \end{tabular}
    \caption{Evaluation of advanced prediction methods for instruction-tuned models  on 4 datasets in 0-shot and 2-shot with random demonstrations. 
    ``Calib.'' stands for the Calibration prompting method.}.
    \label{table:prediction_methods_instruct}
    \end{table*}

\begin{table*}[h!]
    \centering
    \small
    \setlength{\tabcolsep}{1.55pt}
    \begin{tabular}{lccccccccccccc}
    \toprule
    \multirow{2}{*}{Model} & 
    \multicolumn{3}{c}{SST-2} & \multicolumn{3}{c}{DBPedia} & \multicolumn{3}{c}{AGNews} & \multicolumn{3}{c}{TREC} \\
    \cmidrule(lr){2-4}\cmidrule(lr){5-7}\cmidrule(lr){8-10}\cmidrule(lr){11-13}&
    Random & ITM & z-ICL & Random & ITM & z-ICL & Random & ITM & z-ICL & Random & ITM & z-ICL \\
\midrule
Mistral v0.3 7B Instruct & 0.91\textsubscript{0.12} & 0.91\textsubscript{0.10} & 0.86\textsubscript{0.08} & 0.65\textsubscript{0.02} & 0.65\textsubscript{0.06} & 0.37\textsubscript{0.05} & 0.52\textsubscript{0.08} & 0.47\textsubscript{0.10} & 0.37\textsubscript{0.02} & 0.76\textsubscript{0.08} & 0.83\textsubscript{0.06} & 0.67\textsubscript{0.06} \\
Llama 3 8B Instruct & 0.91\textsubscript{0.10} & 0.92\textsubscript{0.08} & 0.77\textsubscript{0.08} & 0.63\textsubscript{0.05} & 0.63\textsubscript{0.08} & 0.26\textsubscript{0.05} & 0.48\textsubscript{0.11} & 0.45\textsubscript{0.09} & 0.26\textsubscript{0.04} & 0.78\textsubscript{0.10} & 0.86\textsubscript{0.04} & 0.53\textsubscript{0.10} \\
\bottomrule
    \end{tabular}
    \caption{Evaluation of advanced selection methods for instruction-tuned models using \textsc{Direct} prediction method in 4-shot. Results are aggregated over 10 random templates for each of the three demonstrations selection seeds.}.
    \label{table:selection_methods_instruct}
    \end{table*}

\subsection{Ensemble Results}
\Cref{table:ensembles_results_instruct} shows that applying Template Ensembles method to instruct models results in improved mean classification accuracy with significantly reduced variance across different templates in all configurations that we test.

\begin{table*}[h!]
\centering
\begin{tabular}{lcccccc}
\toprule
\multirow{2}{*}{Model} & \multicolumn{2}{c}{Direct} & \multicolumn{2}{c}{Channel} & \multicolumn{2}{c}{Calibration} \\
           \cmidrule(lr){2-3}\cmidrule(lr){4-5}\cmidrule(lr){6-7}
     &  Single &      Ensemble &  Single &  Ensemble & Single &   Ensemble \\
\midrule
\multicolumn{7}{c}{SST-2} \\
\midrule
Llama 3 8B Instruct & 0.87\textsubscript{0.13} & 0.94\textsubscript{0.01} & 0.78\textsubscript{0.10} & 0.86\textsubscript{0.02} & 0.90\textsubscript{0.09} & \textbf{0.95}\textsubscript{0.00} \\
Mistral v0.3 7B Instruct & 0.90\textsubscript{0.09} & 0.94\textsubscript{0.01} & 0.83\textsubscript{0.09} & 0.91\textsubscript{0.02} & 0.90\textsubscript{0.12} & \textbf{0.95}\textsubscript{0.00} \\
\midrule
\multicolumn{7}{c}{DBPedia}\\
\midrule
Llama 3 8B Instruct & 0.56\textsubscript{0.12} & 0.64\textsubscript{0.06} & 0.72\textsubscript{0.04} & 0.79\textsubscript{0.01} & 0.85\textsubscript{0.11} & \textbf{0.95}\textsubscript{0.02} \\
Mistral v0.3 7B Instruct & 0.64\textsubscript{0.05} & 0.67\textsubscript{0.02} & 0.75\textsubscript{0.04} & 0.83\textsubscript{0.02} & 0.91\textsubscript{0.07} & \textbf{0.95}\textsubscript{0.05} \\
\midrule
\multicolumn{7}{c}{TREC}\\
\midrule
Llama 3 8B Instruct & 0.39\textsubscript{0.09} & 0.47\textsubscript{0.01} & 0.36\textsubscript{0.09} & 0.43\textsubscript{0.03} & 0.44\textsubscript{0.06} & \textbf{0.49}\textsubscript{0.02} \\
Mistral v0.3 7B Instruct & 0.40\textsubscript{0.08} & \textbf{0.48}\textsubscript{0.02} & 0.41\textsubscript{0.06} & 0.46\textsubscript{0.04} & 0.43\textsubscript{0.06} & 0.47\textsubscript{0.03} \\
\midrule
\multicolumn{7}{c}{AGNews}\\
\midrule
Llama 3 8B Instruct & 0.77\textsubscript{0.18} & \textbf{0.89}\textsubscript{0.02} & 0.72\textsubscript{0.10} & 0.81\textsubscript{0.01} & 0.80\textsubscript{0.08} & 0.84\textsubscript{0.01} \\
Mistral v0.3 7B Instruct & 0.82\textsubscript{0.07} & \textbf{0.86}\textsubscript{0.02} & 0.80\textsubscript{0.02} & 0.83\textsubscript{0.01} & 0.80\textsubscript{0.04} & 0.82\textsubscript{0.02} \\
\bottomrule
\end{tabular}
    \caption{Comparison of 2-shot learning performance of Template Ensembles against other prediction methods on all datasets used in our work. We use ensembles of size 5 and average the results over 5 random seeds}.
    \label{table:ensembles_results_instruct}
    \end{table*}

\end{document}